\documentclass[12pt]{article}
\usepackage{algorithm,algpseudocode,amsmath,amssymb,amsthm,subfig,bm,natbib,tikz,mathtools,placeins,url,fullpage}
\usetikzlibrary{shapes,backgrounds}

\newtheorem{result}{Result}

\DeclareMathOperator*{\argmax}{arg\,max}

\begin{document}

\title{Exact Estimation of Multiple Directed Acyclic Graphs}
\author{Chris. J. Oates, Jim Q. Smith, Sach Mukherjee and James Cussens}
\date{}
\maketitle

\begin{abstract}
\noindent \textbf{Abstract.} This paper considers the problem of estimating the structure of multiple related directed acyclic graph (DAG) models.
Building on recent developments in exact estimation of DAGs using integer linear programming (ILP), we present an ILP approach for joint  estimation over multiple DAGs, 
that does not require that the vertices in each DAG share a common ordering.
Furthermore, we allow also for (potentially unknown) dependency structure between the DAGs. 
Results are presented on both simulated data and fMRI data obtained from multiple subjects.

\smallskip
\noindent \textbf{Keywords.} hierarchical models, directed acyclic graphs, non-exchangeability, integer linear programming, joint estimation

\smallskip
\noindent \textbf{Author footnote.} Chris. J. Oates (E-mail: {\it c.oates@warwick.ac.uk}) and Jim Q. Smith (E-mail: {\it j.q.smith@warwick.ac.uk}), Department of Statistics, University of Warwick, Coventry, CV4 7AL, UK.
Sach Mukherjee (E-mail: {\it sach@mrc-bsu.cam.ac.uk}), MRC Biostatistics Unit and School of Clinical Medicine, University of Cambridge, Cambridge, CB2 0SR, UK.
James Cussens (E-mail: {\it james.cussens@cs.york.ac.uk}), Department of Computer Science and York Centre for Complex Systems Analysis, University of York, York, YO10 5GE, UK.
\end{abstract}

\section{Introduction} \label{intro}

This paper considers joint estimation of multiple directed acyclic graph (DAG) models using integer linear programming (ILP). 
Graphical models are multivariate models 
in which vertices in a graph $G$ represent random variables with  edges between the vertices describing
conditional independence statements concerning the variables. 
In many
settings the edge structure of the graph is itself 
uncertain and then an important challenge is to estimate this structure from
 data.
There has been considerable research into structural inference for graphical
models over the last decade, including Bayesian
networks \citep[BNs;][]{Friedman,Ellis,He}, Gaussian graphical models
\citep[GGMs;][]{Meinshausen,Friedman2008} and discrete graphical
models \citep{Loh}.  
Many graphical models are based on DAGs and these are the focus of this paper.

In many applications, data $\mathcal{D}^{(k)}$ are collected on multiple units $k \in \{1,2,\dots,K\}$ that may differ with respect to conditional independence structure, such that corresponding DAGs $G^{(k)}$ may be non-identical. At the same time, when the units of study are related, the graphs $G^{(k)}$, while potentially non-identical, are expected to have similarities. 
It is then natural to ask whether such similarity can be exploited by borrowing strength across the  estimation problems indexed by $k$ and that is the aim of the present paper.
Specifically we seek to construct a {\it joint} estimator $\hat{G}^{(1:K)}= \hat{G}_{\text{joint}}(\mathcal{D}^{(1)},\dots,\mathcal{D}^{(K)})$ that estimates a collection of DAGs together. We contrast such an estimator with the {\it independent} estimator $\hat{G}^{(k)} = \hat{G}_{\text{indep}}(\mathcal{D}^{(k)})$ that estimates each DAG from the associated data only.

The best known class of DAG models are BNs, but more exotic DAG models exist \citep[e.g.][]{Queen}.
BNs admit an extensive theory of inferred causation that has contributed to their popularity \citep{Pearl}.
Structure learning for individual BNs is a well-studied problem, with contributions including \cite{Friedman,Silander,Tsamardinos,Cowell,Cussens,Jaakola,Yuan}.
Structure learning is NP-hard, but an approach that has attracted much recent attention is to cast {\it maximum a posteriori} (MAP) DAG estimation 
as a problem in integer linear programming (ILP), as
developed simultaneously by \cite{Cussens,Jaakola} and subsequently extended in \cite{Cussens2}.
In brief, this approach solves a sequence of linear relaxations of the MAP estimation problem via the introduction of cutting planes and combines this with
a branch-and-bound search to produce an optimal solution.
If the algorithm terminates, the result is guaranteed to be a global maximum of the posterior distribution and hence inherits theoretical guarantees  associated with the MAP estimator \citep[see e.g.][]{Chickering}.
(In this paper, algorithms with this property are termed ``exact''.) 
Coupled with powerful techniques from discrete optimisation \citep{Nemhauser,Wolsey,Achterberg}, ILPs represent an efficient and attractive methodology for structural inference, as demonstrated also by recent empirical results \cite[e.g.][]{Sheehan}.
Recent advances in this area are discussed in \cite{Bartlett}.

The joint estimation of graphical models has recently received attention, for example \cite{Danaher} put forward a penalised likelihood formulation that couples together estimation for multiple (undirected) GGMs. However, joint estimation of multiple DAGs has so far received relatively little attention.
The first discussion of this problem that we are aware of is \cite{Niculescu}; here a greedy search was used to locate a local maximum of a joint Bayesian posterior. 
\cite{Werhli} described a Markov chain Monte Carlo (MCMC) method for sampling from a joint posterior over graphical structures.
However, the generic difficulties associated with stochastic search/sampling in large discrete spaces are well known; these are exacerbated in the joint case by the size of the joint model space and stochastic search/sampling remains challenging in this setting.

The focus of this paper is instead  on exact, deterministic algorithms.
\cite{Oyen} proposed an exact algorithm based on Bayesian model averaging and belief propagation, under the strong assumption that an ordering of the variables $1,\dots,P$ applies simultaneously to all units.
At the same time \cite{Oates} proposed essentially the same algorithm, applied to the specific class of feed-forward dynamic BNs, where an ordering of the variables is implicitly provided by the time index. 
The algorithmic contributions of the present paper are two-fold. 
First, we show how to cast exact inference over multiple DAGs as an ILP problem.
We consider MAP-Bayesian estimation for multiple DAGs and  require no restriction on the ordering of the variables.
This is done by extending methodology presented in \cite{Bartlett} to the case of multiple DAGs via a  hierarchical Bayesian formulation.
Second, we exploit structural constraints that are imposed by the DAGs in order to improve computational efficiency.
As a illustrative example, our methods currently allow estimation of 10 related DAGs, each with 10 nodes, in time typically less than one minute on a standard laptop.

In addition, we extend previous work by allowing for dependencies between the DAGs themselves and consider also estimation of this dependency structure.
Previous work on multiple DAGs has focused on the special case where the units are exchangeable and all pairs of units undergo an equal amount of regularisation \citep[including][]{Werhli,Oates}.
However, in practice, relationships between units (and their underlying graphical models) may be complex, e.g. hierarchical, with group and sub-group structure, and such structure may itself be subject to uncertainty.
\cite{Oates3} performed exact inference for non-exchangeable feed-forward dynamic BNs for the case where the relationships between units are known {\it a priori}.
\cite{Oyen2} addressed non-exchangeability in the context of general DAG models but did not provide an exact algorithm and, again, assumed that the relationships between units are known at the outset.
Our approach provides a framework that allows for simultaneous learning of both unit-specific DAGs and the dependency structure that relates them. However, our empirical results  suggest that such simultaneous learning may be extremely challenging in practice.

The remainder of the paper is organised as follows:
Section \ref{def the model} introduces a  statistical framework for multiple DAGs and discusses regularisation based on graphical structure.
Section \ref{sec estimate} gives exact, ILP-based estimators for multiple DAGs.
Section \ref{applications} presents a simulation study and results on fMRI data from a multi-subject study.
Finally we close with a discussion of directions for further research.
A companion paper that explores the fMRI application in more detail is available as \cite{Oates5}.

\section{A Statistical Model for Multiple DAGs} \label{def the model}

We begin by introducing the statistical model, deferring  discussion of computation to the next section.
Throughout the shorthand $1:P$ will be used to denote the list of integers $1,2,\dots,P$.
A (directed) graph $G$ on vertices $1:P$ is characterised by a collection of sets $G_i$, such that $G_i \subseteq \{1:P\}\setminus\{i\}$ contains precisely the parents of vertex $i$ according to $G$.
We say $G$ is acyclic if $G$ contains no sequence of directed edges that begins and ends at the same vertex.
Write $\mathcal{G}$ for the space of all directed acyclic graphs (DAGs) with $P$ vertices.
In this paper vertices $i$ in a DAG $G$ will be associated with random variables $Y_i$.
We will use  $Y_i^{(k)}(n)$ to denote the $n$th observation of variable $Y_i$ for unit $k$ and $\bm{Y}_\pi^{(k)}(n)$ to denote the collection $\{Y_i^{(k)}(n) : i \in \pi\}$ of these variables.

\subsection{MAP Estimation}

We present our approach from a MAP-Bayesian perspective, but it could also be  described as a penalised likelihood approach.
We consider DAG models for which the conditional distribution of the variables $\bm{Y}_{1:P}^{(k)} = \bm{Y}_{1:P}^{(k)}(1:N)$, given the DAG $G^{(k)}$ and associated parameters $\bm{\theta}_{1:P}^{(k)} = \bm{\theta}_{1:P}^{(k)}(1:N)$, factorises as
\begin{eqnarray}
p(\bm{Y}_{1:P}^{(k)}|\bm{\theta}_{1:P}^{(k)},G^{(k)}) = \prod_{i=1}^P \prod_{n=1}^N p(Y_i^{(k)}(n)|\bm{Y}_{G_i^{(k)}}^{(k)}(n),\bm{\theta}_i^{(k)}(n),G_i^{(k)}). \label{factor ex}
\end{eqnarray}
Here $G_i^{(k)}$ denotes the parents of the $i$th variable in the DAG $G^{(k)}$ and $\bm{Y}_{G_i^{(k)}}^{(k)}(n)$ are the observed values of these parent variables in sample $n$.
Additionally, $\bm{\theta}_i^{(k)}(n)$ are parameters associated with the conditional distribution for the $i$th variable that may depend on sample index $n$, unit $k$ and model $G_i^{(k)}$, though this latter dependence is suppressed in the notation.
The joint likelihood that we consider below follows from Eqn. \ref{factor ex} and the assumption that the observations $\bm{Y}_{1:P}^{(k)}$ for each unit $k$ are conditionally independent given the DAGs $G^{(k)}$ and associated parameters $\bm{\theta}_{1:P}^{(k)}$. Specifically, we have that the full likelihood factorises over units, variables and samples:
\begin{eqnarray}
p(\bm{Y}_{1:P}^{(1:K)}|\bm{\theta}_{1:P}^{(1:K)},G^{(1:K)}) = \prod_{k=1}^K \prod_{i = 1}^P \prod_{n = 1}^N p(Y_i^{(k)}(n)|\bm{Y}_{G_i^{(k)}}^{(k)}(n),\bm{\theta}_i^{(k)}(n),G_i^{(k)}). \label{likelihood}
\end{eqnarray}
For expositional simplicity, we take the number $N$ of samples to be the same for each unit, though this is not strictly required for our methodology.
Within a Bayesian framework we place a prior distribution $p(\bm{\theta}_i^{(k)}|G_i^{(k)})$ over parameters $\bm{\theta}_i^{(k)}$, such that parameter sets corresponding to units $k$ and $l$ are independent conditional upon the two DAGs $G^{(k)}$ and $G^{(l)}$.
Integrating out the unknown parameters provides the evidence in favour of the joint model $G^{(1:K)} \in \mathcal{G}^K$:
\begin{eqnarray}
p(\bm{Y}_{1:P}^{(1:K)}|G^{(1:K)}) = \int p(\bm{Y}_{1:P}^{(1:K)}|\bm{\theta}_{1:P}^{(1:K)},G^{(1:K)}) p(\bm{\theta}_{1:P}^{(1:K)}|G^{(1:K)}) d \bm{\theta}_{1:P}^{(1:K)}
\end{eqnarray}

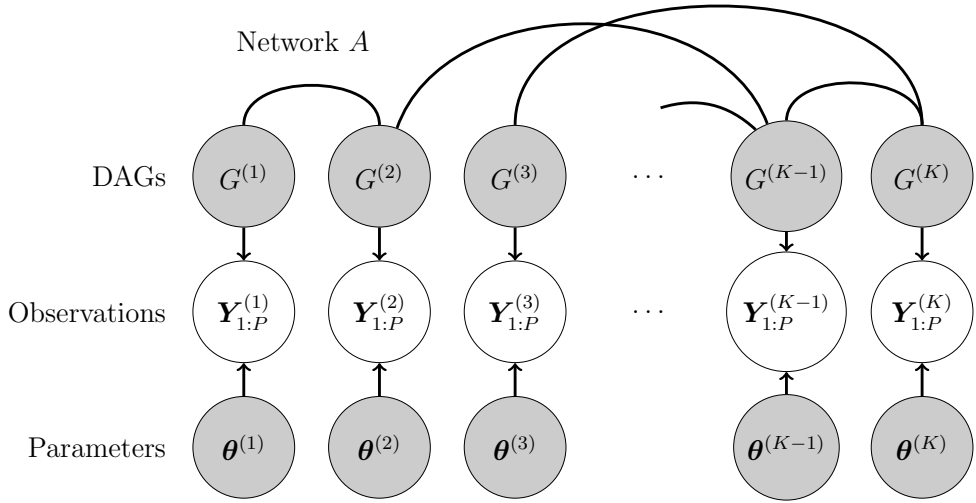
\begin{figure}[t]
\centering
\resizebox{0.8\textwidth}{!}{
\begin{tikzpicture}

\tikzstyle{obs}=[draw,circle,fill = black!0,minimum width=1.5cm];
\tikzstyle{unobs}=[draw,circle,fill = black!20,minimum width=1.5cm];
\tikzstyle{theta}=[draw,circle,fill = black!20,minimum width=1.5cm];
\tikzstyle{arrow}=[very thick,->];
\tikzstyle{line}=[very thick,-];

\node[unobs] at (-4,-2) (G1) {$G^{(1)}$};
\node[unobs] at (-2,-2) (G2) {$G^{(2)}$};
\node[unobs] at (0,-2) (G3) {$G^{(3)}$};
\node[unobs] at (6,-2) (GJ) {$G^{(K)}$};
\node[unobs] at (4,-2) (GJm1) {$G^{(K-1)}$};
\node[obs] at (-4,-4) (y1) {$\bm{Y}_{1:P}^{(1)}$};
\node[obs] at (-2,-4) (y2) {$\bm{Y}_{1:P}^{(2)}$};
\node[obs] at (0,-4) (y3) {$\bm{Y}_{1:P}^{(3)}$};
\node at (2,-2) (dots1) {$\dots$};
\node at (2,-1) (point) {};
\node[obs] at (4,-4) (yJm1) {$\bm{Y}_{1:P}^{(K-1)}$};
\node[obs] at (6,-4) (yJ) {$\bm{Y}_{1:P}^{(K)}$};
\node[theta] at (-4,-6) (t1) {$\bm{\theta}^{(1)}$};
\node[theta] at (-2,-6) (t2) {$\bm{\theta}^{(2)}$};
\node[theta] at (0,-6) (t3) {$\bm{\theta}^{(3)}$};
\node[theta] at (6,-6) (tJ) {$\bm{\theta}^{(K)}$};
\node[theta] at (4,-6) (tJm1) {$\bm{\theta}^{(K-1)}$};
\node at (2,-4) (dots2) {$\dots$};
\path[line] (G1) edge [bend left=90] (G2);
\path[line] (GJm1) edge [bend right] (point);
\path[line] (GJm1) edge [bend left=90] (GJ);
\path[line] (G2) edge [bend left=70] (GJm1);
\path[line] (G3) edge [bend left=90] (GJ);
\path[arrow] (G1) edge (y1);
\path[arrow] (G2) edge (y2);
\path[arrow] (G3) edge (y3);
\path[arrow] (GJm1) edge (yJm1);
\path[arrow] (GJ) edge (yJ);
\path[arrow] (t1) edge (y1);
\path[arrow] (t2) edge (y2);
\path[arrow] (t3) edge (y3);
\path[arrow] (tJm1) edge (yJm1);
\path[arrow] (tJ) edge (yJ);

\node[] at (-2,0) [left] {Network $A$};
\node[] at (-5,-2) [left] {DAGs};
\node[] at (-5,-4) [left] {Observations};
\node[] at (-5,-6) [left] {Parameters};
\end{tikzpicture}
}
\caption{A hierarchical model for multiple directed acyclic graphs (DAGs), with relationships between the DAGs encoded by an undirected network $A$ (shown as edges at top of figure). [Shaded nodes are unobserved. $G^{(1:K)}=$ latent DAGs, $\bm{\theta}^{(1:K)}=$ data-generating parameters, $\bm{Y}_{1:P}^{(1:K)}=$ observations.]}
\label{model}
\end{figure}

Below we introduce our prior distribution over all DAGs $G^{(1:K)} \in \mathcal{G}^K$ that encodes the notion of dependency that we wish to exploit during estimation.
Write $\mathcal{A}$ for the space of undirected networks on vertices $1:K$.
The hierarchical prior that we propose factorises along edges of a network $A \in \mathcal{A}$ whose $K$ vertices correspond to the individual units:
\begin{eqnarray}
p(G^{(1:K)}|A) \propto \left( \prod_{(k,l) \in A} r(G^{(k)},G^{(l)}) \right) \times \left( \prod_{k = 1}^K m(G^{(k)}) \right) \label{joint prior}
\end{eqnarray}
Here the first product ranges over all edges $(k,l)$ in the network $A$.
The (positive, symmetric) function $r(G^{(k)},G^{(l)})$ is interpreted as a measure of regularity (i.e. similarity) between the DAGs $G^{(k)}$ and $G^{(l)}$; specific choices for this function are discussed in Section \ref{choose q} below, motivated by computational convenience in the sequel.
The graph $A$ indicates which pairs of units have similar graphical structure (Fig. \ref{model}).
For example, $A$ may describe a time-ordering of the units, such that consecutive units are expected to have more similar graphical structures, or may indicate group membership within a mixture model.
In most existing literature an exchangeability assumption is placed on $G^{(1:K)}$ that corresponds (implicitly) to specifying $A$ as the complete network \citep[][etc.]{Werhli,Oates,Danaher}, though \cite{Oyen2,Oates3} considered general (known) forms for $A$. 
In this paper we allow for general and potentially unknown $A \in \mathcal{A}$.
The remaining terms $m(G^{(k)})$ are necessary for multiplicity correction and are discussed in Section \ref{multi correct} below, again motivated by computational convenience in the sequel.

Consider the MAP estimator under the multiple DAG prior:
\begin{eqnarray}
\hat{G}^{(1:K)}|A := \argmax_{G^{(1:K)} \in \mathcal{G}^K} p(G^{(1:K)}|\bm{Y}_{1:P}^{(1:K)},A)
\end{eqnarray}
When $A$ is the complete network all pairs of graphs are regularised equally and we refer to this case as ``exchangeable learning''.
More generally, we consider the setting where $A$ itself is unknown.
Then, we impose a prior distribution over $A \in \mathcal{A}$ (described below in Section \ref{A prior}) 
and estimate both the $G^{(k)}$'s and $A$; we refer to this as ``non-exchangeable learning'' and use the MAP estimator
\begin{eqnarray}
(\hat{G}^{(1:K)},\hat{A}) := \argmax_{\substack{G^{(1:K)} \in \mathcal{G}^K \\ A \in \mathcal{A}}} p(G^{(1:K)},A|\bm{Y}_{1:P}^{(1:K)}).
\end{eqnarray}
One of the main contributions of this paper, in Section \ref{sec estimate}, is to prove that for certain choices of the regularity function $r(G^{(k)},G^{(l)})$ discussed below, both $\hat{G}^{(1:K)}|A$ and $(\hat{G}^{(1:K)},\hat{A})$ are characterised as the solutions to ILPs and hence are amenable to exact computation using advanced  techniques such as constraint propagation and cutting plane algorithms.

\subsection{A Default Choice of Regularity Function} \label{choose q}

Below we discuss choices for the regularity function $r(G^{(k)},G^{(l)})$ that forms the basis for the multiple DAG prior and captures similarity between DAGs $G^{(k)}$ and $G^{(l)}$.
Write $\stackrel{+C}{=}$ for equality up to an unspecified additive constant.
The default choice that we consider for $r$ is
\begin{eqnarray}
\log(r(G^{(k)},G^{(l)})) \stackrel{+C}{=} - \sum_{i=1}^P \sum_{j=1}^P \lambda_{j,i}^{(k,l)} [ (j \in G_i^{(k)}) \oplus (j \in G_i^{(l)}) ]. \label{hm}
\end{eqnarray}
Here $\oplus$ is the logical XOR operator, $[\cdot] \in \{0,1\}$ is an indicator associated with its (logical) argument and the $\lambda_{j,i}^{(k,l)}$ are (constant) penalty terms associated with the inclusion of the edge $(j,i)$ in exactly one of the DAGs $G^{(k)}$ and $G^{(l)}$.
Combined with the default multiplicity correction discussed below, the prior defined by Eqn. \ref{hm} is hyper-Markov with respect to any DAG $G^{(k)}$ when $G^{(l)}$ is held fixed \citep{Dawid}; we will see in Section \ref{sec estimate} that such priors permit a particularly simple construction of an ILP for the multiple DAG model.
For many scientific applications where graph structure represents a physical mechanism, Eqn. \ref{hm} can often be motivated from physical considerations. 
For example in neuroscience, an edge in a graphical model may have the interpretation of physical or functional connectivity and transfer of information between regions of the brain \citep[so-called ``reification''; e.g.][]{Costa}.

A special case, that treats both units and edges as exchangeable, is the structural Hamming distance (SHD) obtained by setting $\lambda_{j,i}^{(k,l)} = \lambda \in [0,\infty)$ $\forall i,j,k,l$ in Eqn. \ref{hm}.
SHD has previously been used to regularise between graphical models by \cite{Niculescu,Penfold,Oyen,Oates} and to integrate prior knowledge into BNs by \cite{Acid,Tsamardinos,Perrier,Hill}.
An extension of SHD with two degrees of freedom was also considered in this context by \cite{Werhli}.
The empirical results presented in this paper focus on SHD due to its interpretability and simplicity, but the methodology is compatible with the general form Eqn. \ref{hm} and the $\lambda_{j,i}^{(k,l)}$'s could be used to encode prior information on the similarity between units or the propensity for the presence of a particular edge to be conserved between units.

\subsection{A Default Choice of Multiplicity Correction} \label{multi correct}

The function $m(G)$ in the multiple DAG prior (Eqn. \ref{joint prior}) is required to adjust for the fact that the size of the space $\mathcal{G}$ grows super-exponentially with the number $P$ of vertices \citep{Consonni}.
We follow \cite{Scott} and control multiplicity using the default binomial correction
\begin{eqnarray}
m(G) = \prod_{i = 1}^P m_i(G_i) , \; \; \; \; \; m_i(\pi) = \binom{P}{|\pi|}^{-1} [ |\pi| \leq d_{\max} ].
\end{eqnarray}
Here $d_{\max}$ is an upper bound on the (maximum) in-degree of $G$ (such upper bounds are widely used to control the computational intensity of structural inference). 
This specification has the desirable property that the collective prior probability of all models with $d$ predictors is $(1+d_{\max})^{-1}$, which is independent of both $d$ and $P$.

We note that the methods presented in this paper are compatible with the inclusion of additional terms in Eqn. \ref{joint prior} that 
encode specific informative priors on the DAGs $G^{(k)}$ but we do not use such priors here.

\subsection{A Default Choice of Hyper-prior} \label{A prior}

We use the following prior for the undirected graph $A$ that encodes pairwise relationships between the units $1 \ldots K$:
\begin{eqnarray}
\log(p(A)) \stackrel{+C}{=} \sum_{k=1}^K \sum_{l=k+1}^K \eta^{(k,l)} [ (k,l) \in A ].
\end{eqnarray}
\noindent 
(We adopt the convention that the adjacency matrix for   $A$ is upper triangular, hence $k < l$ above.)
Here the $\eta^{(k,l)}$'s encourage inclusion of the corresponding edges $(k,l)$ in $A$ and could in principle be used to encode specific knowledge regarding similarity between the units, although we do not pursue this direction here. In all empirical results below we consider the simplest case of  
$\forall k,l \; \eta^{(k,l)} = \eta \in [0,\infty)$. Then, $\eta$ can be viewed as an inverse temperature hyper-parameter, with larger values of $\eta$ encouraging denser networks, which in turn correspond to a greater amount of between-unit regularisation.


\subsection{Elicitation of Hyper-parameters}

All joint estimators for graphical models that we are aware of require tuning parameters \citep[e.g. the ``fused'' and ``group'' flavours of the joint graphical lasso each have two tuning parameters;][]{Danaher}.
Recalling that the DAGs $G^{(1:K)}$ give a probability model for the observations $\bm{Y}_{1:P}^{(1:K)}$, hyper-parameters in the priors above may be set using  standard procedures such as cross-validation or information criteria.
Let $\bm{\phi}$ denote all hyper-parameters to be set.
We recommend the use of information criteria for their computational convenience.
In particular, 
using the Akaike information criteria (AIC) we have
\begin{eqnarray}
\hat{\bm{\phi}} = {\arg\max}_{\bm{\phi}} \;  \log p(\bm{Y}_{1:P}^{(1:K)}|\hat{G}^{(1:K)}(\bm{\phi})) - \sum_{k=1}^K \sum_{i=1}^P \dim(\bm{\theta}_i^{(k)}|\hat{G}_i^{(k)}(\bm{\phi})) \label{AIC}
\end{eqnarray}
where $\dim(\bm{\theta}_i^{(k)}|G_i^{(k)})$ are the number of parameters required to specify the conditional distribution of $\bm{Y}_i^{(k)}$ given $\bm{Y}_{G_i^{(k)}}^{(k)}$.
Thus the AIC requires only that we can obtain the MAP estimate $\hat{G}^{(1:K)}(\bm{\phi})$ over a range of values of the hyper-parameters $\bm{\phi}$, selecting the value that maximises Eqn. \ref{AIC}.
In contrast, cross-validation requires the marginal predictive likelihood function $p(\bm{Y}_{1:P}^* | \bm{Y}_{1:P}^{(1:K)})$ for held-out data $\bm{Y}_{1:P}^*$, which for many models may require nontrivial additional computation.
We note that a grid search is necessary in the case of multiple tuning parameters; sequential optimisation is not possible due to the non-orthogonality induced by the joint prior $p(G^{(1:K)},A)$. For more complex joint priors/penalties this represents a challenge that we do not address here.

\section{Integer Linear Programs For Joint Estimation} \label{sec estimate}

We now consider the computational aspects of MAP estimation for the models introduced above. We begin with the simpler case where the network $A$ is known {\it a priori} and subsequently consider the more general unknown $A$ case.

\subsection{Exact Estimation of Multiple DAGs When $A$ is Known} \label{fix N}

The methodology we present below extends the ILP formulation of \cite{Jaakola} to multiple units via the inclusion of additional state variables that capture the similarities and differences between the units.
We begin by computing and caching the terms
\begin{eqnarray}
p(\bm{Y}_i^{(k)}|\bm{Y}_{G_i^{(k)}}^{(k)},G_i^{(k)}) = \int p(\bm{Y}_i^{(k)}|\bm{Y}_{G_i^{(k)}}^{(k)},\bm{\theta}_i^{(k)},G_i^{(k)})p(\bm{\theta}_i^{(k)} | G_i^{(k)}) d\bm{\theta}_i^{(k)} \label{evidence}
\end{eqnarray}
that summarise evidence in the data for the local model $G_i^{(k)}$ for the $i$th variable in unit $k$.
These are available in closed form for many models of interest, including but not limited to discrete BNs with Dirichlet priors \citep{Heckerman}, linear Gaussian structural equation models with conjugate priors \citep{Pearl} and multiregression dynamical models \citep{Queen}.
Since our approach is based upon these pre-computed quantities, they could even be obtained numerically (for more complex models) using MCMC and related techniques \citep[e.g.][]{Oates6,Oates7}.
 These cached quantities are transformed to obtain ``local scores'', defined as
\begin{eqnarray}
s^{(k)}(i,G_i^{(k)}) := \log(p(\bm{Y}_i^{(k)}|\bm{Y}_{G_i^{(k)}}^{(k)},G_i^{(k)})) + \log(m_i(G_i^{(k)})).
\end{eqnarray}
\noindent These are the (log-) evidence from Eqn. \ref{evidence} with an additional penalty term that provides multiplicity correction over varying $G_i^{(k)} \subseteq \{1:P\} \setminus \{i\}$.

We define binary indicator variables $[ G_i^{(k)} = \pi ]$ that form the basis of our ILP. 
Here $\pi \subseteq \{1:P\}\setminus\{i\}$ is used to denote a possible parent set for the $i$th variable in the DAG $G^{(k)}$, so the information in the variables $[ G_i^{(k)} = \pi ]$ completely characterises $G^{(k)}$.  
It will be necessary to impose constraints that ensure these variables correspond to a well-defined DAG:
\begin{equation}
\sum_{\pi \subseteq \{1:P\} \setminus \{i\}} [ G_i^{(k)} = \pi ] = 1  \; \; \; \; \;  \forall i,k\tag{C1; convexity}
\end{equation}
Constraint (C1) requires that for each unit $k$, every vertex $i$ has exactly one parent set (i.e. there is a well-defined graph $G^{(k)}$).
To ensure $G^{(k)}$ is acyclic we require further constraints:
\begin{equation}
\sum_{i \in C} \sum_{\substack{\pi \subseteq \{1:P\}\setminus\{i\} \\ \pi \cap C = \emptyset}} [ G_i^{(k)} = \pi ] \geq 1 \; \; \; \; \;  \forall k, \emptyset \neq C \subseteq \{1: P\}. \tag{C2; acyclicity}
\end{equation}
(C2) states that for every non-empty set $C$ there must be at least one vertex in $C$ that does not have a parent in $C$.
It is not challenging to prove (by contradiction) that (C1-2) exactly characterise the space $\mathcal{G}$ of DAGs.

Next we define an indicator $[j \in G_i^{(k)}]$ of the presence of each specific edge $(j,i)$ in $G^{(k)}$.
This can be related to the parent set indicators as
\begin{equation}
[j \in G_i^{(k)}] = \sum_{\substack{\pi \subseteq \{1:P\}\setminus\{i\} \\ j \in \pi}} [G_i^{(k)} = \pi] \; \; \; \; \;  \forall i,j,k \tag{C3} .
\end{equation}
To decide whether units $k,l$ agree on the presence or absence of a specific edge $(j,i)$, we introduce additional variables $\delta^{(k,l)}(j,i) := [ (j \in G_i^{(k)}) \oplus (j \in G_i^{(l)}) ]$ where $\oplus$ is the logical XOR operator.
To encode this definition we use the set of linear inequalities
\begin{align}
+ \delta^{(k,l)}(j,i) && - [j \in G_i^{(k)}] && -[j \in G_i^{(k)}] && \leq &&
0 \label{xor1} \tag{C4.1} \\
- \delta^{(k,l)}(j,i) && +[j \in G_i^{(k)}] && -[j \in G_i^{(k)}] && \leq && 0 \tag{C4.2} \\
- \delta^{(k,l)}(j,i) && -[j \in G_i^{(k)}] && +[j \in G_i^{(k)}] && \leq && 0 \tag{C4.3} \\
+ \delta^{(k,l)}(j,i) && +[j \in G_i^{(k)}] && +[j \in G_i^{(k)}] && \leq && 2 \tag{C4.4}
\end{align}
These inequalities are optimal in the sense that they define (in the most concise way) the convex hull (in $\mathbb{R}^3$) of feasible solutions to the corresponding 3-variable XOR constraint \citep{Achterberg2}.
Write $\bm{x}$ for a vector containing the binary variables $[G_i^{(k)} = \pi]$, $[j \in G_i^{(k)}]$ and $\delta^{(k,l)}(j,i)$, so that $\bm{x} \in \mathcal{X} = \{0,1\}^q$ where $q = O(KP2^P) + O(KP^2) + O(K^2P^2)$.
We can now state our first result:
\begin{result} \label{fixed N}
The MAP estimate $\hat{G}^{(1:K)}|A$ is characterised as the solution of the ILP
\begin{eqnarray}
\hat{G}^{(1:K)}|A = \argmax_{\bm{x} \in \mathcal{X}} \sum_{k=1}^K \sum_{i=1}^P \sum_{\pi \subseteq \{1:P\}\setminus\{i\}} s^{(k)}(i,\pi) [G_i^{(k)} = \pi] -  \sum_{(k,l) \in A} \sum_{i = 1}^P \sum_{j=1}^P \lambda_{j,i}^{(k,l)} \delta^{(k,l)}(j,i)  \label{IP}
\end{eqnarray}
subject to constraints (C1-4).
\end{result}
\noindent Note that under an in-degree restriction $|\pi| \leq d_{\max}$ the dimensionality of the state space $\mathcal{X}$ can be reduced from exponential in $P$ to polynomial in $P$, specifically $q = O(KP^{1+d_{\max}}) + O(KP^2) + O(K^2P^2)$.
It is seen that the dimensionality of $\mathcal{X}$ increases by a factor $O(K)$ when learning $K$ DAGs jointly, as opposed to individually (assuming that the first term dominates).

\begin{figure}
\centering
\includegraphics[width = 0.6\textwidth,clip,trim = 3cm 19.5cm 9cm 2.5cm]{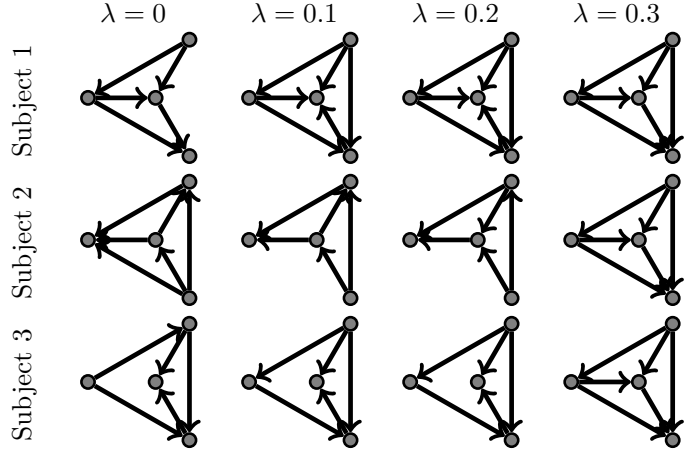}
\caption{Illustration of exchangeable learning on a small fMRI dataset  consisting of measurements of activity at 4 brain regions (nodes) in 3 subjects (rows).
Structure learning was performed using the proposed model for multiple DAGs with regularity parameter $\lambda \in \{0,0.1,0.2,0.3\}$ (columns).
For increasing $\lambda$ the DAGs become progressively more similar.
MAP estimates $\hat{G}^{(1:3)}|A$ are shown with $A$ set equal to the complete network.
}
\label{example}
\end{figure}

To illustrate the use of Result \ref{fixed N}, consider a fixed network $A$ that is equal to the complete network (exchangeable learning) and regularisation based on SHD with hyper-parameter $\lambda$.
Fig. \ref{example} displays the MAP $\hat{G}^{(1:K)}|A$ computed (exactly) as a function of the regularity hyper-parameter $\lambda$ for a small fMRI dataset of 3 subjects containing measurements of activity in 4 neural regions.
The first row, with $\lambda = 0$, is the result of performing independent estimation on the 3 subjects to obtain MAP estimators for subject-specific DAGs.
Notice that the DAGs are quite dissimilar; the neuroscience context from which these data arise suggests that this is most likely an artefact (i.e. due to variance in estimation).
As the regularity parameter $\lambda$ is increased, the subject-specific DAGs become more similar until they are eventually identical at $\lambda = 0.3$ and above.
In simulation studies below we demonstrate that this kind of exchangeable learning can lead to improved estimation of unit-specific graphical structure.

\subsection{Exact Estimation of Multiple DAGs When $A$ is Unknown} \label{unknown N}

The previous section addressed the case where the interdependency structure $A$ among the units was known.
Below we formulate an ILP for the more general case where $A$ is unknown and must be estimated jointly with the unit-specific DAGs.

\subsubsection{Exact Estimation of $G^{(1:K)}$ and $A$} \label{learn N}

We introduce additional binary indicator variables $[ (k,l) \in A ]$ that directly encode those edges that are present in the network $A$.
Taking together the variables $\delta^{(k,l)}(j,i)$ and $[ (k,l) \in A ]$ allows us to determine whether or not an edge $(j,i)$ differs between an unit $k$ and its neighbour $l$ in $A$.
Specifically, we represent this information by defining a new binary variable $\Delta^{(k,l)}(j,i) := [ ((j \in G_i^{(k)}) \oplus (j \in G_i^{(l)})) \& ((k,l) \in A) ]$ where $\&$ is the logical AND operator.
To encode this definition in an ILP we use the following set of linear inequalities
\begin{align}
+ \Delta^{(k,l)}(j,i) && -[(k,l) \in A] &&  && \leq && 0 \tag{C5.1} \\
+ \Delta^{(k,l)}(j,i) &&  && -\delta^{(k,l)}(j,i) && \leq && 0 \tag{C5.2} \\
- \Delta^{(k,l)}(j,i) && +[(k,l) \in A] && +\delta^{(k,l)}(j,i) && \leq && 1\tag{C5.3}.
\end{align}
The linear inequalities are again optimal, in the sense that they define (in the most concise way) the convex hull (in $\mathbb{R}^3$) of feasible solutions \citep{Achterberg2}.
In this case we write $\bm{x}$ for a binary vector that additionally contains the variables $[(k,l) \in A]$ and $\Delta^{(k,l)}(j,i)$, so that $\bm{x} \in \mathcal{X} = \{0,1\}^q$ with $q$ larger than in the exchangeable setting, but with the same asymptotic scaling.
We can now state our second result:
\begin{result} \label{nonex theo}
The MAP estimate $(\hat{G}^{(1:K)},\hat{A})$ is characterised as the solution of the ILP
\begin{eqnarray}
(\hat{G}^{(1:K)},\hat{A}) := \argmax_{\bm{x} \in \mathcal{X}} && \sum_{k=1}^K \sum_{i=1}^P \sum_{\pi \subseteq \{1:P\}\setminus\{i\}} s^{(k)}(i,\pi) [G_i^{(k)} = \pi] \label{IP2}  \\
& & - \sum_{i=1}^P \sum_{j=1}^P \sum_{k=1}^K \sum_{l=k+1}^K  \lambda_{j,i}^{(k,l)} \Delta^{(k,l)}(j,i) + \sum_{k = 1}^K \sum_{l=k+1}^K \eta^{(k,l)} [(k,l) \in A] \nonumber
\end{eqnarray}
subject to constraints (C1-5).
\end{result}
\noindent The dimensionality of $\mathcal{X}$ can again be reduced from exponential in $P$ to polynomial in $P$ through the adoption of an in-degree restriction.

\subsection{An Efficient Computational Implementation} 

Our computational implementation is built on the C package GOBNILP that performs inference for individual DAGs \citep{Bartlett}, available at \url{http://www.cs.york.ac.uk/aig/sw/gobnilp/}.
GOBNILP, originally described in \cite{Cussens2}, computes the MAP estimate on a per-unit basis; we therefore created an interface that interacts with GOBNILP and enforces additional constraints (C3-5) that couple together multiple DAGs as described above.
Below we elaborate on how such an interface was constructed and summarise additional optimisation routines that underpin associated computational efficiency in the joint setting.

Since there are exponentially many acyclicity constraints (C2),
GOBNILP does \emph{not} construct an integer program (IP) containing
all of them. Instead, initially, only the convexity constraints (C1)
are included in an IP. GOBNILP then (exactly) solves the linear program (LP) relaxation of the IP where integer variables are (temporarily) allowed to take any real
value within their bounds. 
This produces an LP solution
$\bm{x}^*$. Crucially, the LP solution can be
found very quickly. GOBNILP then searches for (C2) acyclicity
constraints that $\bm{x}^*$ violates and adds them to the IP. Linear
constraints added in this way are known as ``cutting planes'' since
they `cut off' the infeasible solution $\bm{x}^*$.  GOBNILP then solves the
linear relaxation of this new IP and looks for cutting planes for the
new LP solution. This process of solving linear relaxations and adding
cutting planes is then repeated. Note that the sequence of LP
solutions produced in this way provide increasingly tight upper bounds
on the objective value of the MAP estimator.

It can happen that an LP solution represents well-defined DAGs (i.e.\ all
variables have integer values and the graphs they represent are each
acyclic). In this case, the problem is solved exactly, since any LP
solution provides an upper bound on the objective function. More typically (at least on larger
problems) a fractional LP solution is generated that satisfies all
the constraints. In such a situation GOBNILP selects a
fractional variable, e.g. $[G_i^{(k)} = \pi]$, to branch on, creating two
subproblems, one where $[G_i^{(k)} = \pi]=0$ (ruling out parent set $\pi$
for vertex $i$ for unit $k$) and one where $[G_i^{(k)} = \pi]=1$ (setting the parent
set of vertex $i$ for unit $k$ to be $\pi$). The GOBNILP algorithm is then
applied recursively on both branches. When an IP variable
$[G_i^{(k)} = \pi]$ is set to 1, GOBNILP performs constraint propagation by
setting to 0 those IP variables that indicate other parent sets for vertex $i$
or parent set choices for other vertices in the DAG that are no longer possible under the acyclicity constraints.

This approach to IP solving is known as ``branch-and-cut'' since both
branching on variables and adding cutting planes are used. GOBNILP is
implemented using the SCIP framework due to \cite{Achterberg}. The basic GOBNILP algorithm described
above has been optimised in a number of ways: e.g.\ extra linear constraints
are added to the initial IP and additional cutting plane algorithms
(some built into SCIP, some GOBNILP-specific) are used. In addition, a
heuristic algorithm (based on `rounding' the current LP solution) is
used to generate `good' but probably sub-optimal DAGs. This can help
prune the search tree produced by branching on variables (but does not affect exactness of the algorithm). See
\citet{Bartlett} for further details. 

For joint estimation we constructed an interface for GOBNILP that interlaces with SCIP to include additional constraints that couple together estimation problems for multiple DAGs.
In practice (C4) are implemented using SCIP's XOR constraint
handler. This handles any constraint of the form $r = x_{1} \oplus
x_{2} \oplus \dots \oplus x_{n}$ where $r,x_i \in \{0,1\}$ and the constraint is satisfied if either
$r=1$ and an odd number of the $x_i$ are 1, or $r=0$ and an even number of the $x_i$ are 0.  We
set $0 = \delta^{(k,l)}(j,i) \oplus [j \in G_i^{(k)}] \oplus
[j \in G_i^{(l)}]$. 
SCIP generates these inequalities internally and also provides a propagator for
XOR constraints; in the simplest case, as soon as any two of $\delta^{(k,l)}(j,i)$,
$[j \in G_i^{(k)}]$ and $[j \in G_i^{(l)}]$ have their values fixed then SCIP
will immediately fixed the value of the third appropriately.
Similarly (C5) are implemented using SCIP's AND constraint
handler. This handles any constraint of the form $r = x_{1} \&
x_{2} \& \dots \& x_{n}$, where $r,x_i \in \{0,1\}$. We set  $\Delta^{(k,l)}(j,i) = [(k,l) \in A] \&
\delta^{(k,l)}(j,i)$. 
SCIP provides a propagator for AND constraints that works analagously as for XOR constraints.

A naive implementation of our methodology can be quite effective, but with some additional consideration we can reduce computational effort by about an order of magnitude.
Specifically, we introduce additional (redundant) constraints
\begin{equation}
[j \in G_i^{(k)}] + \sum_{\substack{\pi \subseteq \{1:P\}\setminus\{i\} \\ j \notin \pi}} [G_i^{(k)} = \pi] = 1 \; \; \; \; \;  \forall i,j,k. \tag{C6; efficiency} 
\end{equation}
These do not change the definition of the ILP, since intuitively (C6) states that if the edge $(j, i)$ is present in $G^{(k)}$ then the parent set $G_i^{(k)}$ for vertex $i$ must contain the vertex $j$, which is trivially true.
Nevertheless the inclusion of (C6) can greatly reduce search time in practice since it allows for more effective propagation.
This is because when (C6) are included then the edge-indicators $[j \in G_i^{(k)}]$ become particularly good variables to branch on, producing a more balanced and consequently smaller search tree.

In this work we restricted attention to the MAP estimator, but the GOBNILP software and our interface allow efficient computation of the top $n$ solutions to the ILP, that is the $n$ best joint estimates of the DAGs, along with their associated values for the objective function (and hence Bayes factors).
The interfacing software is written for MATLAB R2014b and is provided in the supplementary materials.

\section{Applications} \label{applications}
In this Section we investigate the performance of the proposed joint estimators, using simulated data and data from a neuroscience study. In all examples, we restrict attention to a single regularity hyper-parameter $\lambda$ and a single density hyper-parameter $\eta$.

\subsection{Simulation Study} \label{sec sim}

\begin{figure}[t!]
\centering
\subfloat[Exchangeable Learning]{
\begin{tabular}{|r|ccc|} \hline
& $P=4$ & 8 & 12 \\ \hline
$K=4$ &  232/4 & 1,488/14 & 4,536/503 \\
8 & 656/5  & 3,872/419 & 11,184/$*$  \\
12 & 1,272/10 & 7,152/7,992  & 19,944/$*$ \\ \hline
\end{tabular}}
\subfloat[Non-exchangeable Learning]{
\begin{tabular}{|r|ccc|} \hline
& $P=4$ & 8 & 12 \\ \hline
$K=4$ & 310/1 & 1,830/59 & 5,334/89 \\
8 & 1,020/9 & 5,468/30,112 & 14,908/$*$ \\
12 & 2,130/17 & 10,914/193,047 & 28,722/$*$ \\ \hline
\end{tabular}}
\caption{Worst-case analysis: Length $q$ of the binary state vector \citep[after pre-solving;][]{Mahajan} / solving time (in seconds), for a single ILP with $P$ vertices and $K$ units.
[Here an in-degree restriction $d_{\max} = 2$ was used and local scores $s^{(k)}(i,\pi)$ were generated independently from $N(0,1)$, leading to very many DAG models attaining similar posterior probabilities.
An asterisk $*$ is used to indicate an ``out of memory'' error, due to an intractable number of branches in the search tree.
All simulations were performed on a single core CPU @ 3GHz with 3877MB RAM.]}
\label{times}
\end{figure}

The performance of structure learning algorithms for graphical models can be quantified in many ways \citep[e.g.][]{Peters2,Oates8}.
Structural estimation can be viewed as  a binary classification task on individual edges and performance can be measured by the Matthews correlation coefficient (MCC) which is popular in scenarios where the number of negative samples (i.e. non-edges) can out-weigh the number of positive samples (i.e. edges).

All experiments below were repeated 10 times based on independently generated datasets.
In order to generate multiple related DAGs we employed a MCMC sampling scheme that targets the multiple DAG prior $p(G^{(1:K)}|A)$. 
Metropolis-Hastings proposals that switch the status (i.e. present or absent) of an edge (selected uniformly at random) within a unit (selected uniformly at random), subject to DAG constraints, achieved an average acceptance probability of 83\%.
Then, conditional on a DAG $G^{(k)}$, we simulated data $\bm{Y}_{1:P}^{(k)}$.
This could be achieved using essentially any likelihood model and in principle this choice will affect the properties of our estimators.
We therefore took the simplest approach of directly simulating the evidence terms (Eqn. \ref{evidence}) from log-normal distributions.
Full pseudocode and convergence diagnostics are provided in the supplementary materials.

We emphasise that the multiple rounds of estimation performed in the simulation study below preclude comparison with more computationally intensive approaches such as \cite{Werhli}.
On the other hand, methods that require a shared ordering of the variables such as \cite{Oyen,Oates3} do not apply in this general setting.
In all experiments below we select hyper-parameters using a brute-force grid search over the (coarse) grid $\lambda \in \{0,\frac{1}{2},1,2,\infty\}$, $\eta \in \{0,\frac{1}{2},1,2,\infty\}$.
That is, we solve up to 25 ILPs for each evaluation of the AIC-based estimator.
These candidate values for hyper-parameters were deliberately chosen to match the scale of the simulated data, as explained in the supplementary material.
In practice the combinations of hyper-parameters that yield poor estimates (i.e. low AIC) tend to correspond to ILPs that are quickly solved, so that this grid search does not substantially affect computational complexity.
To establish an (approximate) upper bound on the computational time required to evaluate our proposed estimators, we generated scores $s^{(k)}(i,\pi)$ independently from $N(0,1)$, leading to very many DAG models attaining similar posterior probabilities.
This is a ``worst case scenario'' for the ILP solver, in the sense that maximum effort is required to distinguish between the models.
An indication of these upper bounds on the computational times for a single ILP solve is provided in Fig. \ref{times}.
It is seen that the operating range of our procedure is restricted to approximately $P,K \leq 10$, though in practice the worst-case scenario is unlikely to be realised and it may be possible to solve larger problems (this is highly dependent on the values taken by the local scores).
We make two further observations:
Firstly, in the exchangeable case with $P=12$, $K=8$ the program was terminated due to insufficient RAM.
However the current best solution up to that point achieved an objective value of 71.49 and the upper bound provided by the ILP was 93.31.
This ability to provide a sure upper bound enables a heuristic MAP search for more challenging problems (though in the sequel we considered only exact solutions).
Secondly, although the non-exchangeable case with $P=8$, $K=12$ required 193,047 seconds to terminate (about 54 hours), the MAP estimate was actually found after 31,020 seconds (16\% of the total time), with the remaining time used to confirm that this was indeed the MAP.
This suggests that computational time could be reduced if we are willing to sacrifice the guarantee of exactness.

\begin{figure}[t!]
\centering
\subfloat[Regime (i); $P = 5$, $K = 4$, $\lambda_{\text{true}}=0.65$]{\includegraphics[width = 0.49\textwidth,clip,trim = 0cm 0cm 0cm 0.7cm]{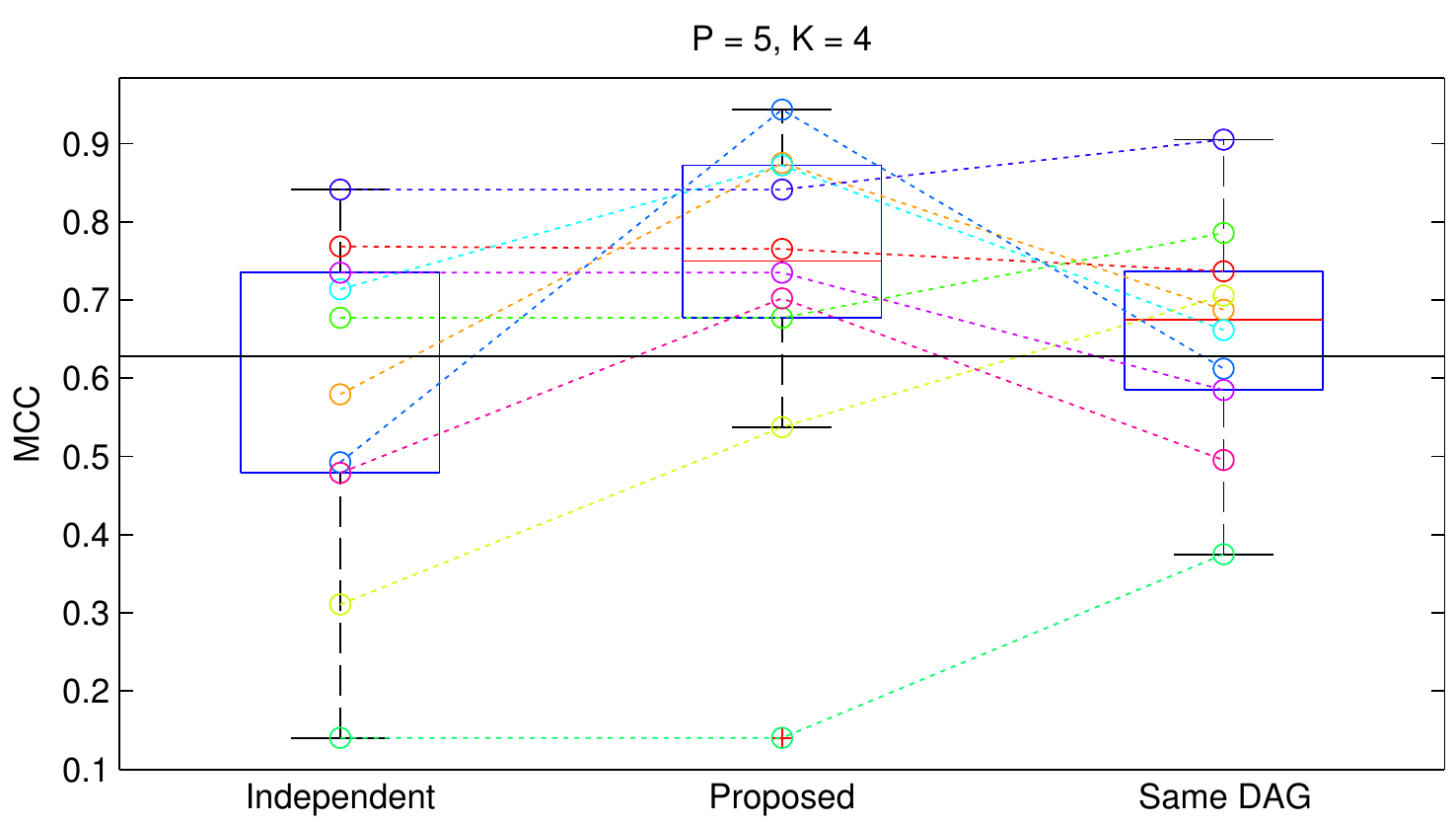}}
\subfloat[Regime (i); $P = 5$, $K = 8$, $\lambda_{\text{true}}=0.3$]{\includegraphics[width = 0.49\textwidth,clip,trim = 0cm 0cm 0cm 0.7cm]{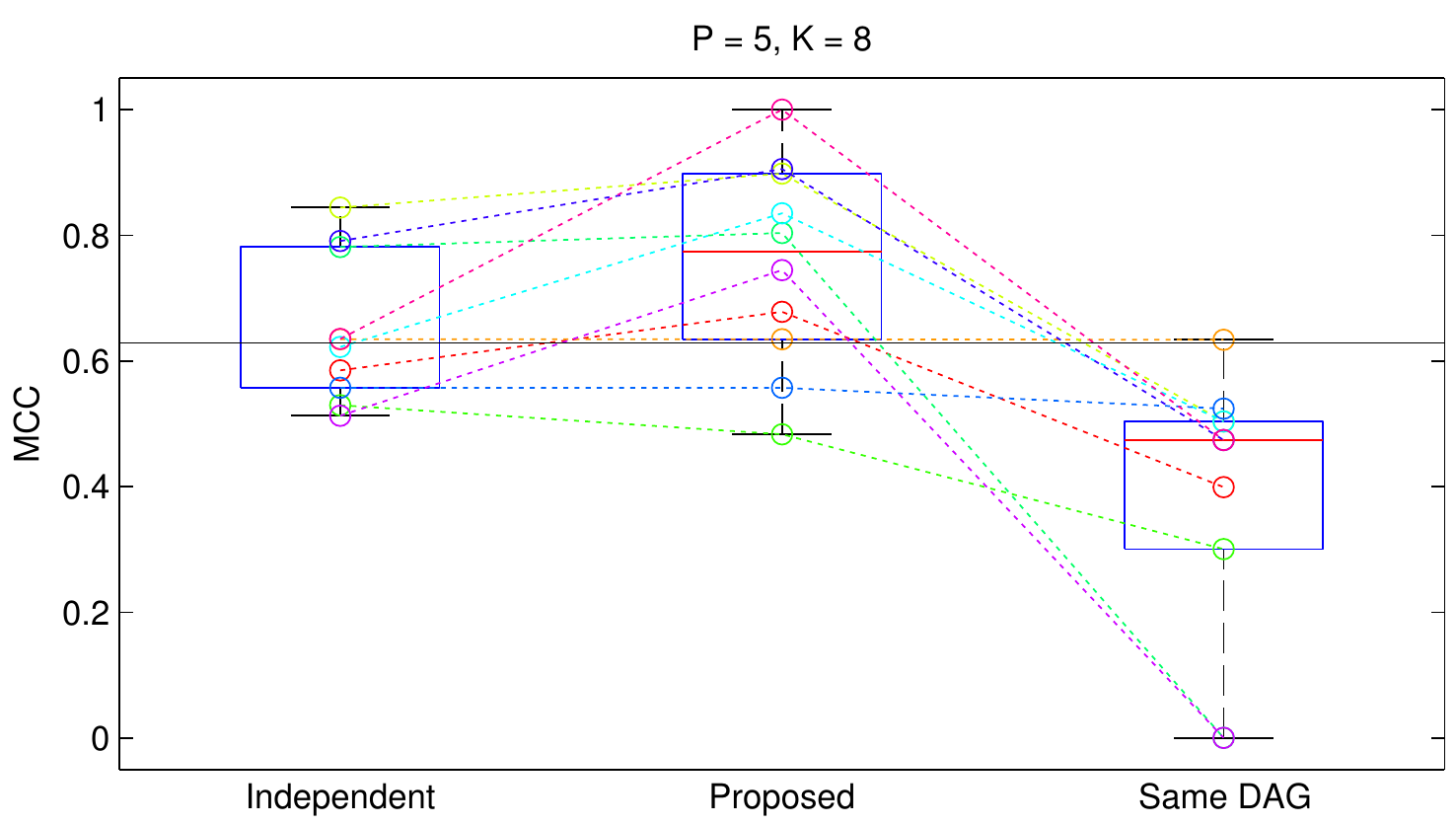}}

\subfloat[Regime (i); $P = 10$, $K = 4$, $\lambda_{\text{true}}=0.65$]{\includegraphics[width = 0.49\textwidth,clip,trim = 0cm 0cm 0cm 0.7cm]{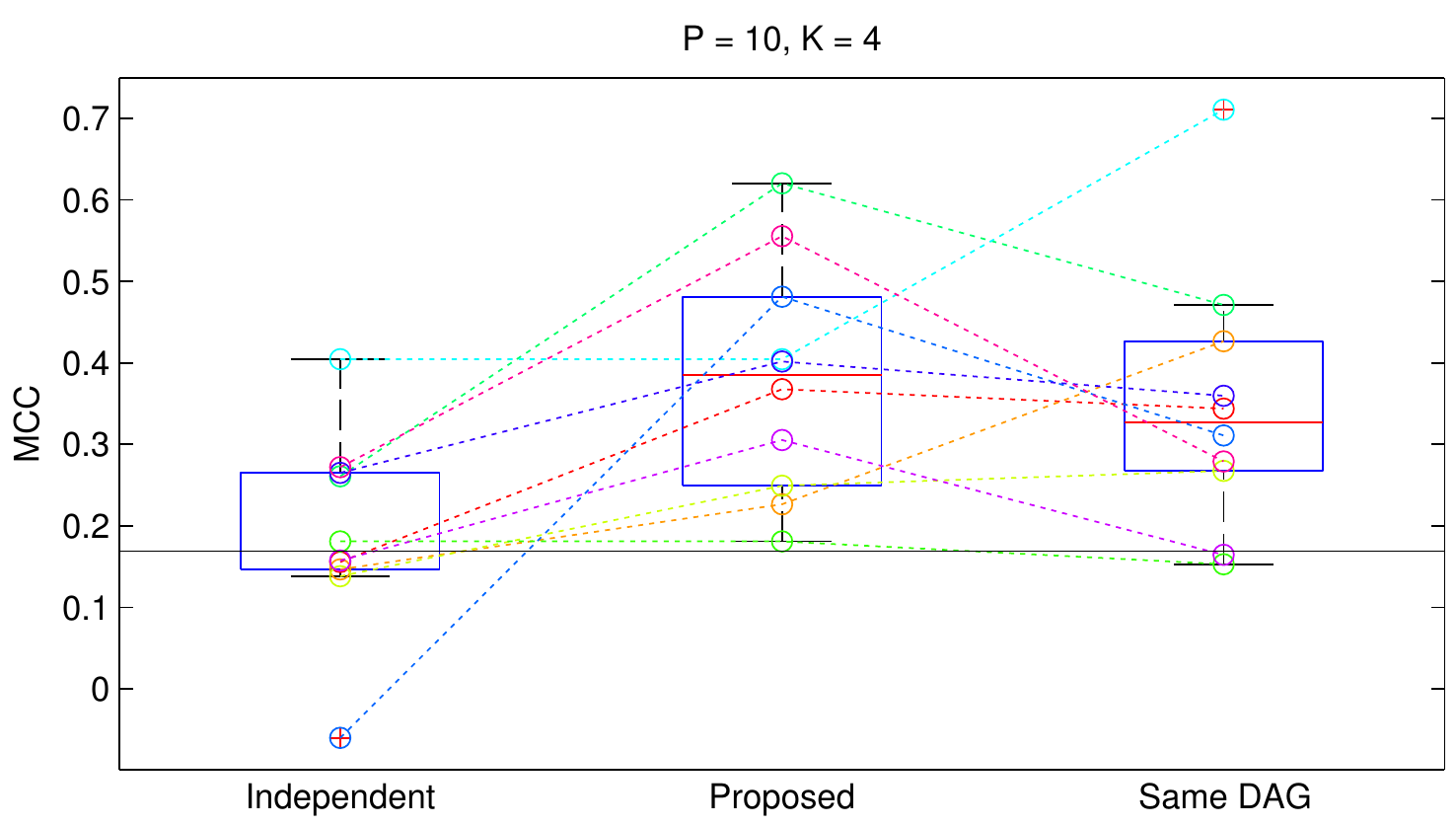}}
\subfloat[Regime (i); $P = 10$, $K = 8$, $\lambda_{\text{true}}=0.3$]{\includegraphics[width = 0.49\textwidth,clip,trim = 0cm 0cm 0cm 0.7cm]{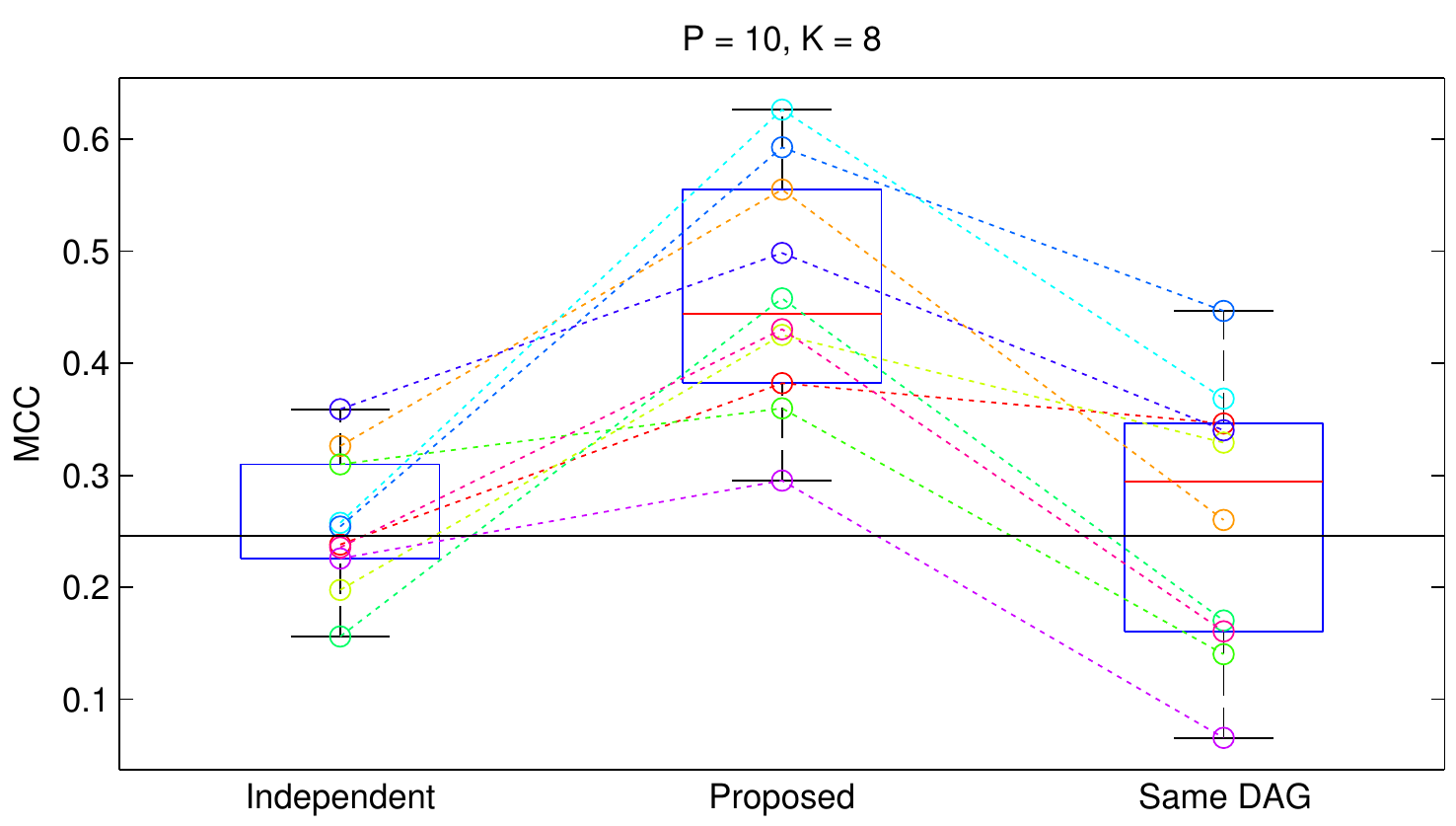}}

\subfloat[Regime (ii); $P = 5$, $K = 4$, $\lambda_{\text{true}}=0.65$]{\includegraphics[width = 0.49\textwidth,clip,trim = 0cm 0cm 0cm 0.7cm]{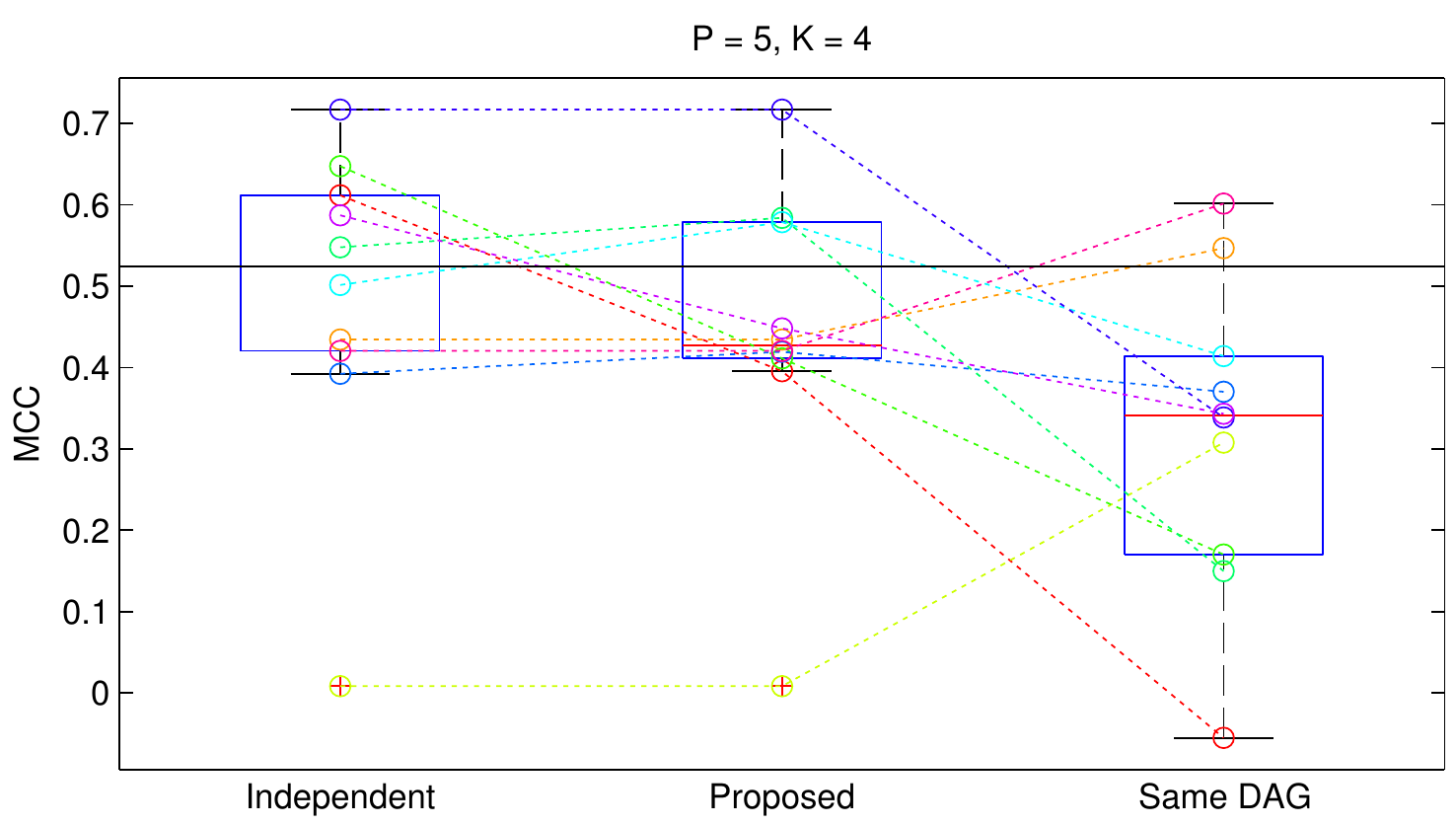}}
\subfloat[Regime (ii); $P = 5$, $K = 8$, $\lambda_{\text{true}}=0.3$]{\includegraphics[width = 0.49\textwidth,clip,trim = 0cm 0cm 0cm 0.7cm]{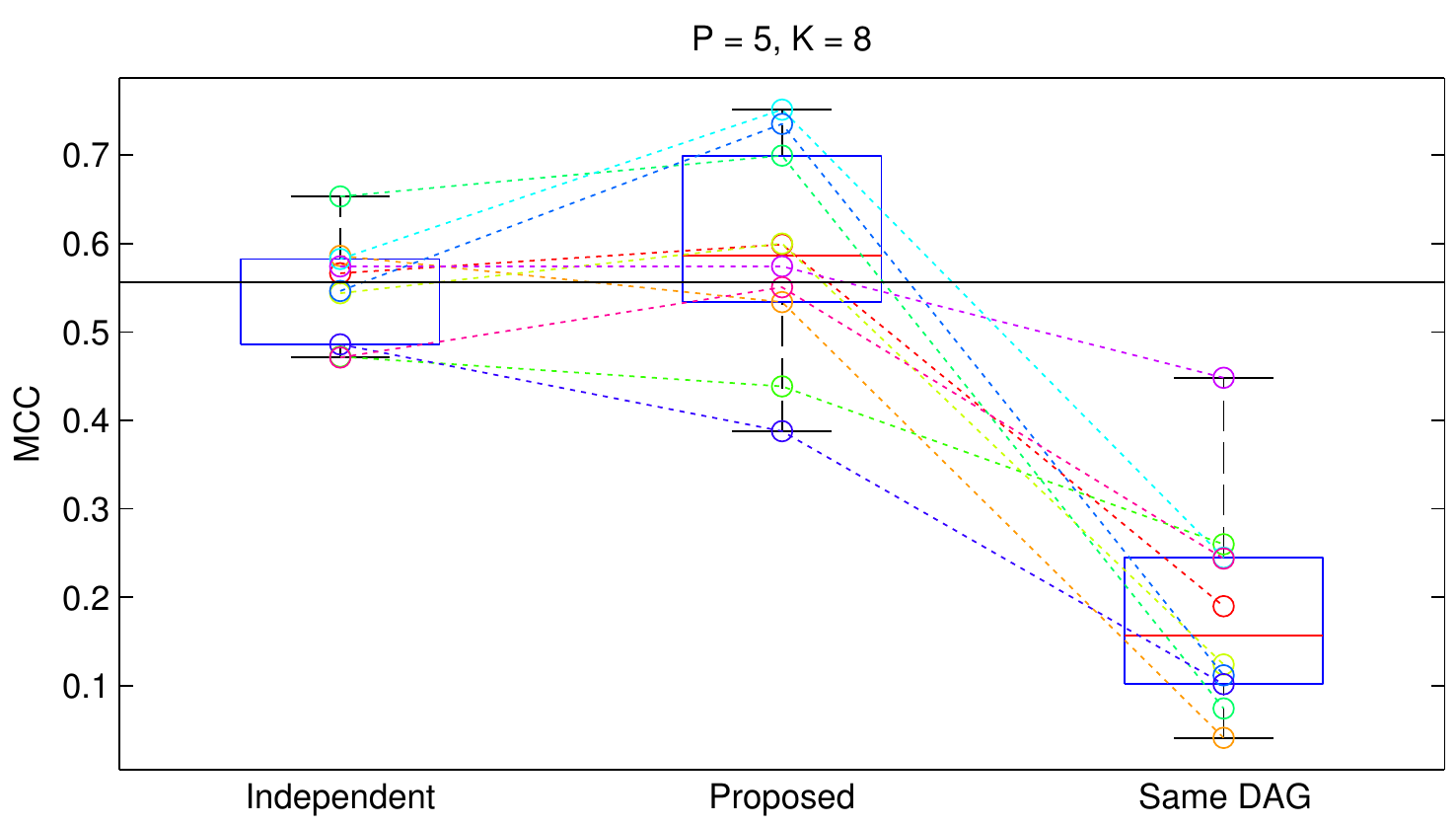}}

\caption{Exchangeable learning of multiple DAGs.
[In regime (i) the data-generating DAGs satisfy an exchangeability assumption.
In regime (ii) the data-generating DAGs strongly violate exchangeability (see main text for details).
Here $P$ is the number of variables, $K$ is the number of units, MCC is the Matthews correlation coefficient and $\lambda_{\text{true}}$ is the data-generating regularity parameter.
Horizontal lines are the median MCC achieved under independent inference.
For fair comparison the same 10 datasets were subject to estimation using each of the three approaches; these are shown as dashed lines.]}
\label{sims1}
\end{figure}

{\bf Exchangeable learning:} Firstly we considered the case where the network $A$ (that describes relationships between the units $k$) is fixed equal to the complete network (exchangeable learning). We focussed on two distinct data-generating regimes: (i) The data-generating process has $A$ complete with related but non-identical DAGs, and (ii) the data-generating process has $A$ as two disconnected complete components of equal size and therefore DAGs more similar within each component than between components. 
Regime (i) employs exchangeable learning in a favourable setting where information is shared between all DAGs, whilst regime (ii) employs exchangeable learning in an unfavourable setting, where the assumption of exchangeability is violated.
Results for regime (i) in Fig. \ref{sims1} (a-d) demonstrate that substantially improved performance is achieved by the proposed joint estimators compared with independent estimation. 
We also compared against the pooled estimator that requires all DAGs to have identical structure - a crude form of information sharing.
Results showed that this ``Same DAG'' estimator performed worse in general than inference based on hyper-parameters selected using AIC, as expected.
The same datasets were employed for each estimator, indicated by dashed lines, to ensure a fair comparison.
Results from regime (ii) in Fig. \ref{sims1} (e-f) show that, despite the exchangeability assumption being violated, exchangeable learning remains competitive with independent estimation, but ``Same DAG" performs poorly, as expected.

\begin{figure}[t!]
\centering
\subfloat[Regime (i); $P = 5$, $K = 4$, $\lambda_{\text{true}}=0.65$]{\includegraphics[width = 0.49\textwidth,clip,trim = 0cm 0cm 0cm 0.7cm]{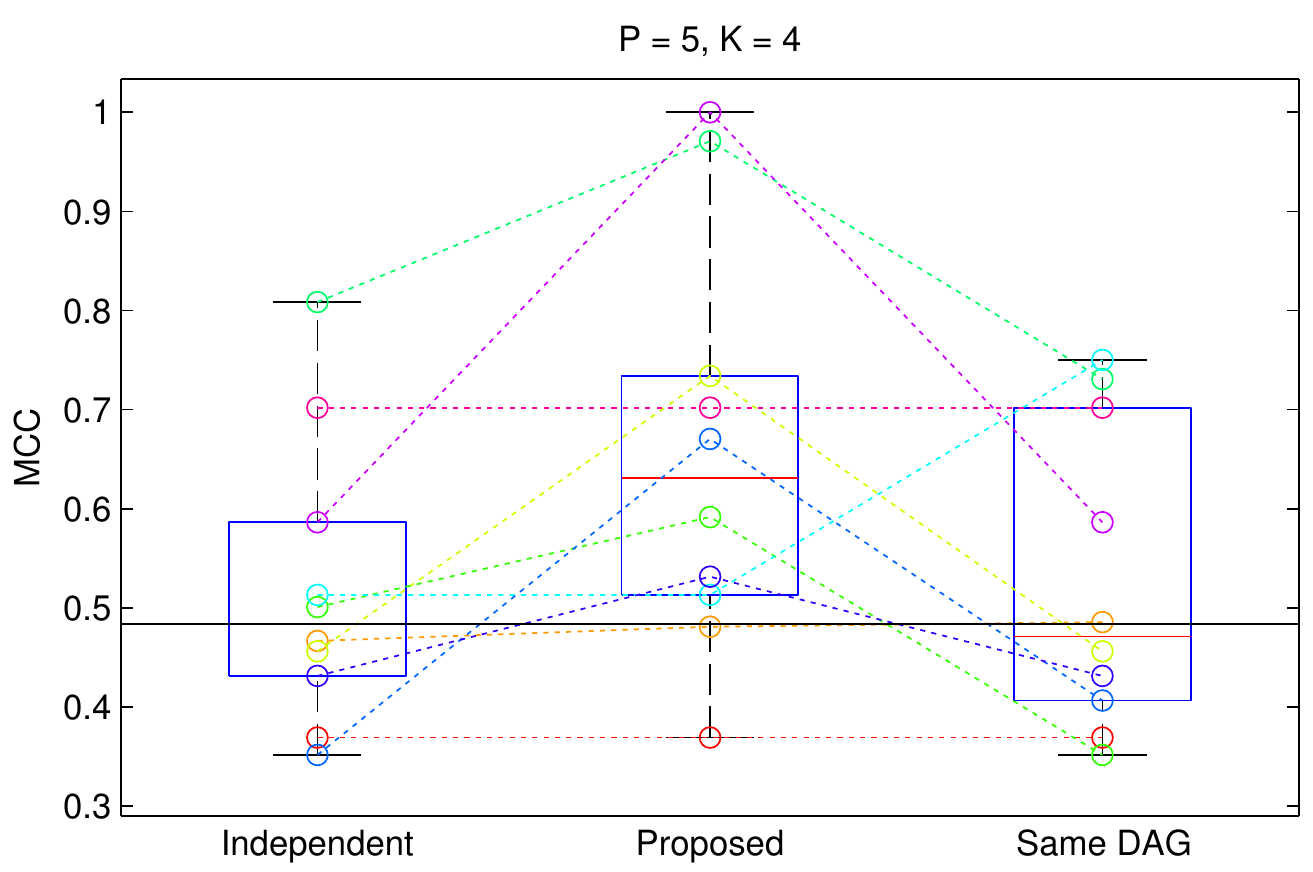}}
\subfloat[Regime (i); $P = 5$, $K = 8$, $\lambda_{\text{true}}=0.3$]{\includegraphics[width = 0.49\textwidth,clip,trim = 0cm 0cm 0cm 0.7cm]{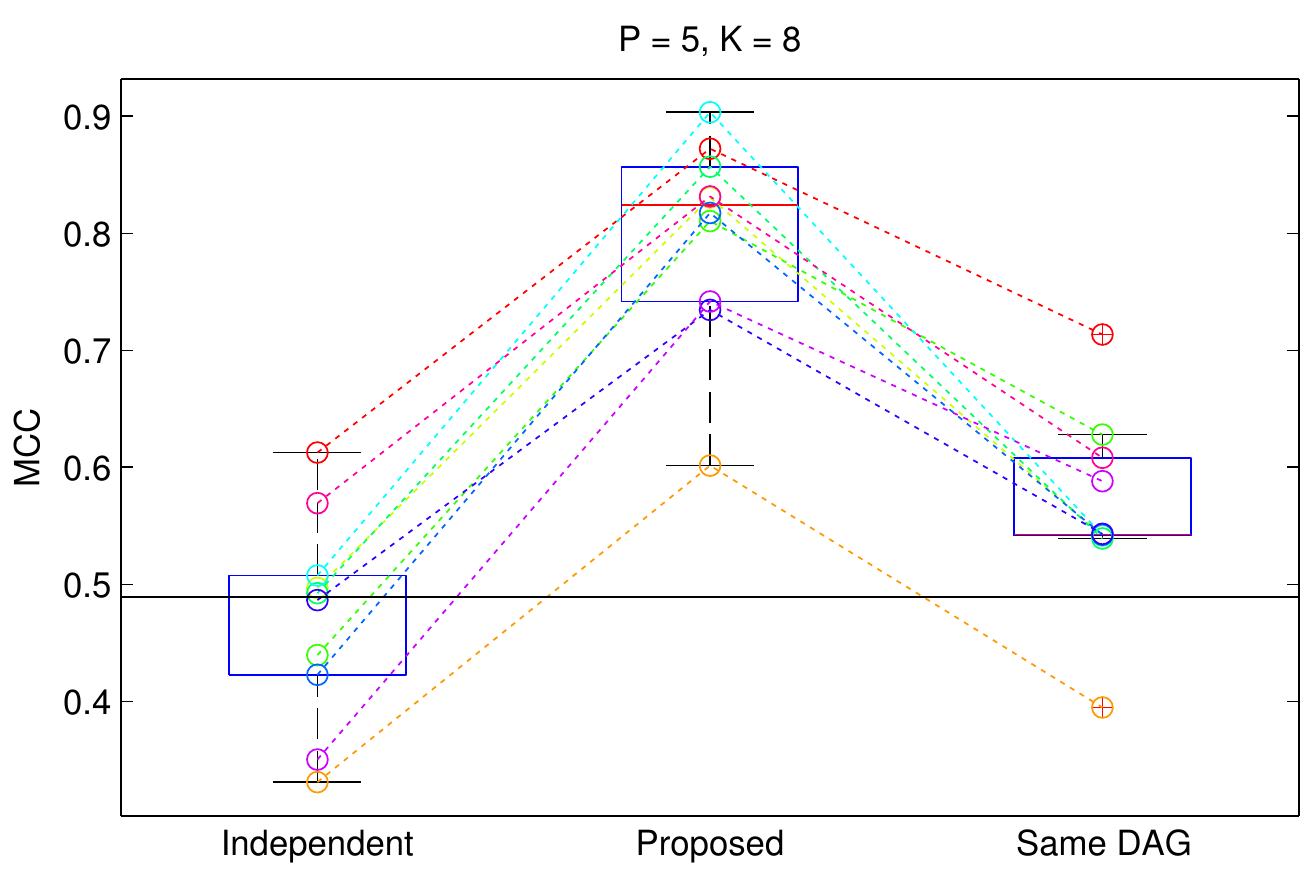}}

\subfloat[Regime (ii); $P = 5$, $K = 4$, $\lambda_{\text{true}}=0.65$]{\includegraphics[width = 0.49\textwidth,clip,trim = 0cm 0cm 0cm 0.7cm]{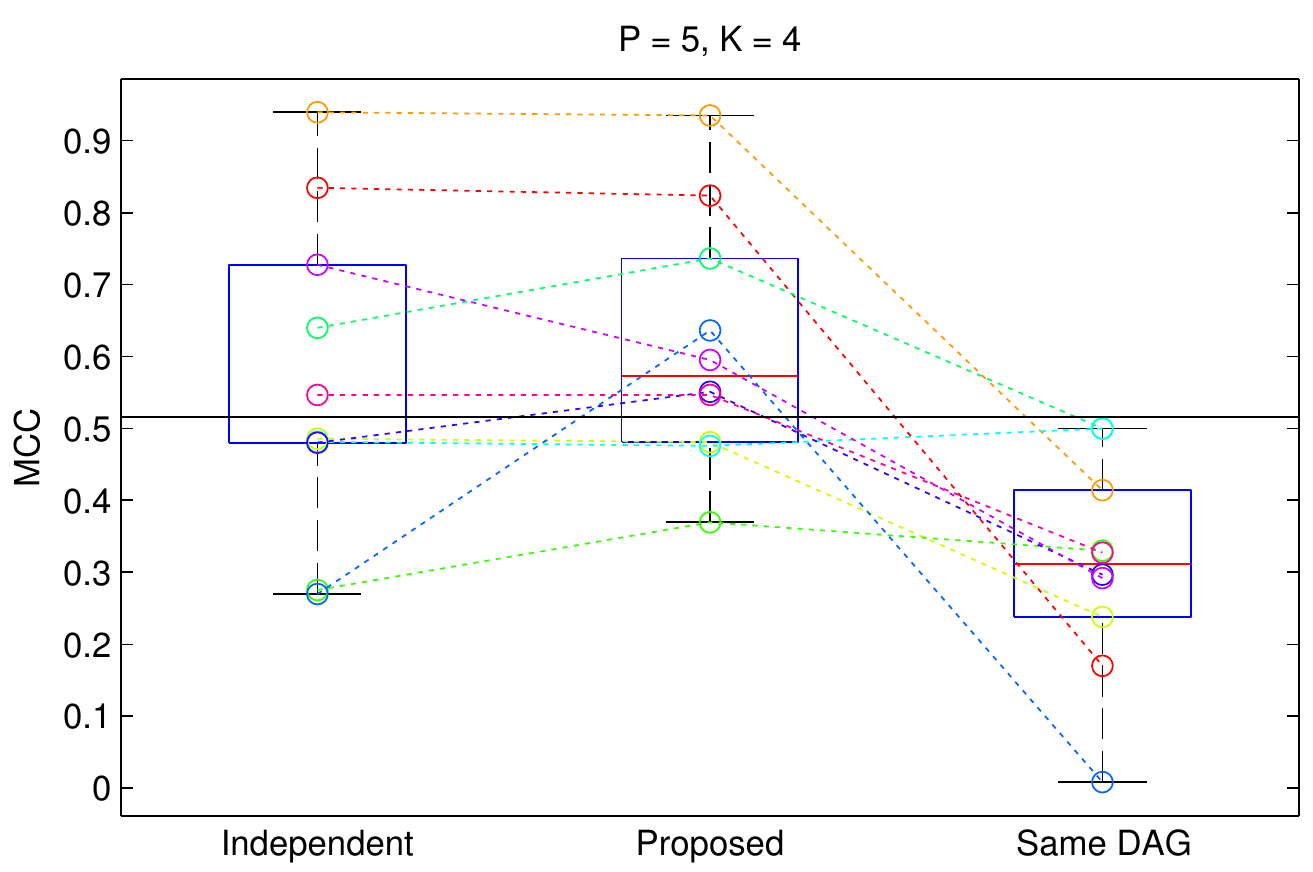}}
\subfloat[Regime (ii); $P = 5$, $K = 8$, $\lambda_{\text{true}}=0.3$]{\includegraphics[width = 0.49\textwidth,clip,trim = 0cm 0cm 0cm 0.7cm]{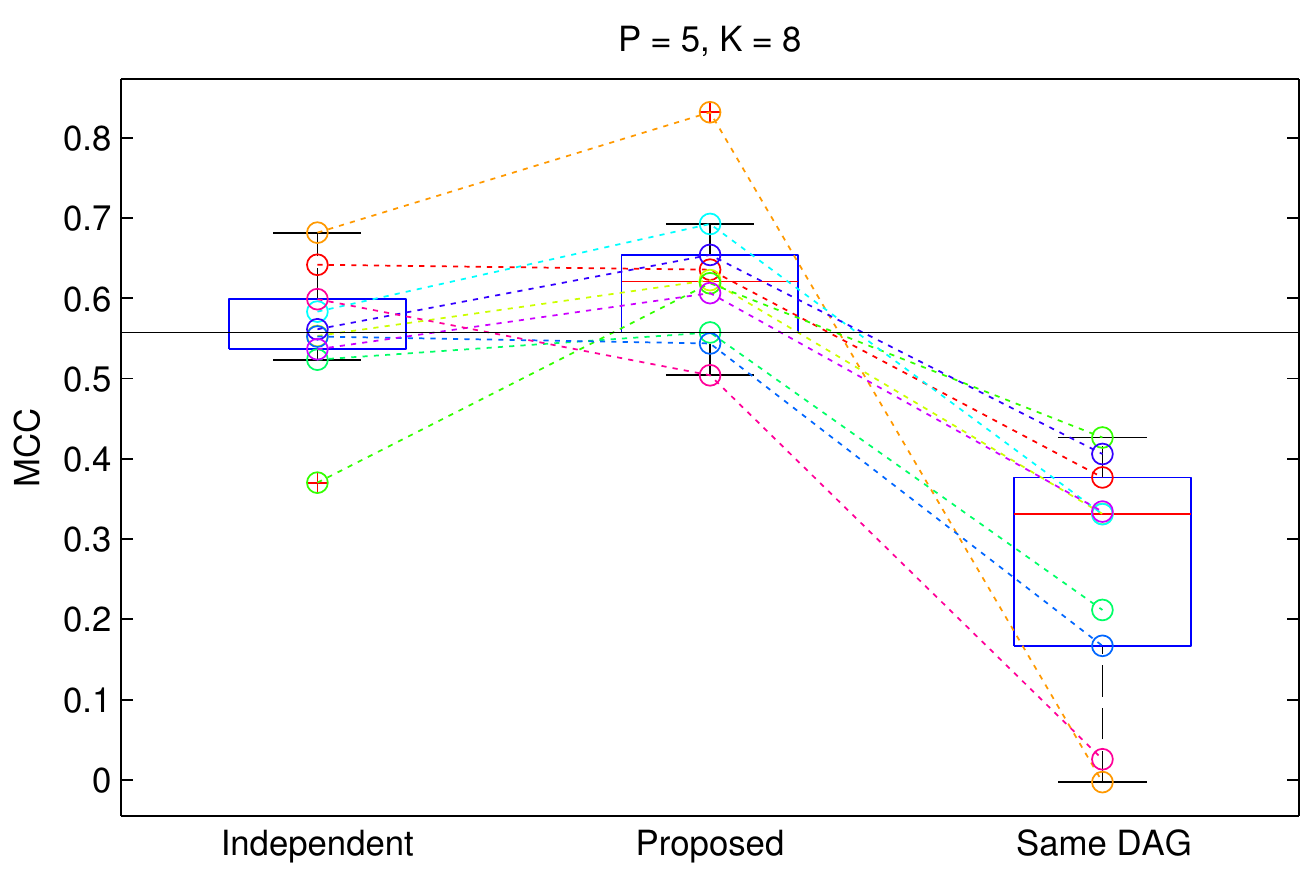}}
\caption{Non-exchangeable learning of multiple DAGs.
[In regime (i) the data-generating DAGs satisfy an exchangeability assumption, whereas in regime (ii) the data-generating DAGs violate exchangeability (see main text for details).
Here $P$ is the number of variables, $K$ is the number of units, MCC is the Matthews correlation coefficient and $\lambda_{\text{true}}$ is the data-generating regularity parameter.
Horizontal lines are the median MCC achieved under independent inference.
For fair comparison the same 10 datasets were subject to estimation using each of the three approaches; these are shown as dashed lines.]}
\label{sims3}
\end{figure}

{\bf Non-exchangeable learning:} Secondly we considered the case where the network $A$ is unknown and must be learned along with the unit-specific DAGs $G^{(1:K)}$. 
In general the data-generating $A$ could have any topology; we again focussed on the two distinct regimes described above.
Regime (i) aims to examine the loss of statistical efficiency that results from employing non-exchangeable learning when exchangeable learning would be more suitable, whilst regime (ii) explores the challenging case where the unknown $A$ is highly informative, so that the loss of information that comes from not knowing $A$ is greatest.
Results for regime (i) in Fig. \ref{sims3}(a-b) demonstrate that, despite being extremely general, non-exchangeable learning still achieves improved estimation compared to both independent inference and the Same DAG estimator.
For regime (ii), shown in Fig. \ref{sims3}(c-d), we see that estimation is more challenging.
Indeed the gains in performance compared with independent estimation were small, perhaps reflecting the increased difficulty of jointly learning both $G^{(1:K)}$ and $A$.
(As before, the Same DAG procedure performed poorly in regime (ii).)

\begin{figure}[t!]
\centering
\includegraphics[width = 0.49\textwidth]{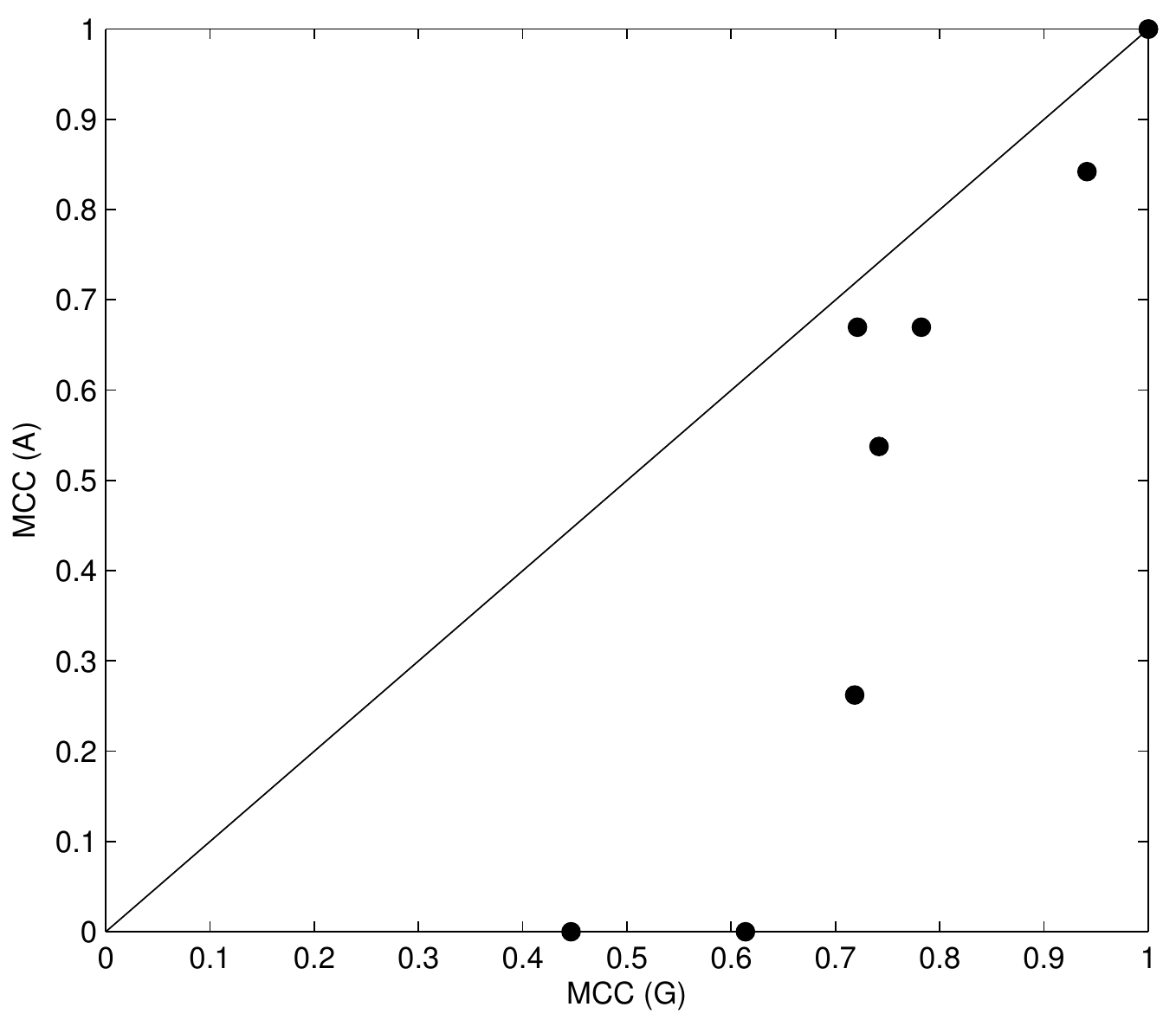}
\caption{Comparing estimation of unit-specific DAGs $G^{(1:K)}$ with estimation of the network $A$ that describes relationships between units.
[Data were generated from regime (ii),where the DAGs violate exchangeability (see main text for details), with $P=5$, $K = 8$.
For clarity we fixed hyper-parameters $\lambda$, $\eta$ to values appropriate for the data-generating process.
MCC is the Matthews correlation coefficient, computed for both the DAGs $G^{(1:K)}$ and the network $A$.]}
\label{sim5}
\end{figure}

For regime (ii) it is interesting to ask whether the network $A$ is itself accurately estimated; results indicated that $A$ is not well estimated, with the MAP network $\hat{A}$ typically being equal to the complete network.
Results in Fig. \ref{sim5} compare the ability to estimate $G^{(1:K)}$ with the ability to estimate $A$, showing that $A$ was typically more difficult to estimate compared with the unit-specific DAGs $G^{(1:K)}$.
This may explain why we do not see substantive gains over independent estimation in the non-exchangeable case, with performance similar to exchangeable learning.
The difficulty in estimating $A$ is perhaps unsurprising: $A$ is latent for the DAGs $G^{(1:K)}$ that are themselves latent, so that  the ``doubly latent'' $A$ may be only weakly identifiable.
Thus, while DAG estimation results in the unknown $A$ formulation 
seem at worst competitive with the simpler exchangeable set-up, 
we would recommend caution  in interpreting the estimate $\hat{A}$ itself.

\subsection{MDMs for Neural Activity in a Multi-Subject Study} \label{neuro app}

We illustrate the methodology by analysing data  from a functional magnetic resonance imaging (fMRI) experiment.
Graphical models are widely used to represent neural connectivity in such studies and it is typical for fMRI experiments to involve multiple subjects.
A statistical analysis should therefore take into account not only the interaction between areas of one brain but also the similarities and differences among subjects \citep{Mechelli}. 
For instance, using fMRI data derived from a writing experiment \citet{Sugihara} showed that the communication pattern among brain regions may be subject-specific, while \citet{Li} argue that the causal relations between brain regions can vary in disease states according to the severity of the disease. 
We analysed a small fMRI dataset consisting of six unrelated subjects from the Human Connectome Project \citep{VanEssen}.  
Data were acquired on each subject while they were in a state of quiet repose and preprocessed to obtain 10-dimensional time series representing the activity levels at 10 regions in each subject.
The specific application is discussed in greater detail in the companion paper \cite{Oates5}.

The interactions (or connections) among different cerebral areas are usually studied through causal DAG models \citep{Friston,Poldrack}. 
Here we use the multiregression dynamical model (MDM) for fMRI data developed by \cite{Costa}, in which edges denote direct contemporaneous relationships that might exist between nodes and the connections are represented by parameters that vary over time.
Thus MDMs are a generalisation of classical ``static'' BNs but, unlike BNs, are fully identifiable (i.e. the equivalence classes are singletons). 
The MDM is defined on a multivariate time series that aims to identify the causal structure among the variables over time \citep{Queen,Queen2}. In the MDM that we consider, a multivariate model for observable series $\bm{Y}_{1:P}^{(k)}(n)$, for subject $k$ at time $n$ is characterised by a contemporaneous DAG, with information shared across time only through evolution of the model parameters $\bm{\theta}_i^{(k)}(n)$.
Formally, this model is described by the following observation equations and system equations:  
\begin{eqnarray}
Y_i^{(k)}(n) & = & \mathbf{Y}_{G_i^{(k)}}^{(k)}(n)^T \boldsymbol{\theta}_i^{(k)}(n) + \epsilon_i^{(k)}(n) , \; \; \; \; \;  \epsilon_i^{(k)}(n) \sim N(0,V_ i^{(k)}(n)) \label{mdmeq1} \\
\boldsymbol{\theta}^{(k)}(n) & = & \bm{\Gamma}^{(k)}(n) \boldsymbol{\theta}^{(k)} (n-1) + \mathbf{w}^{(k)}(n) , \; \; \; \; \; \mathbf{w}^{(k)}(n)\sim N (\mathbf{0},\mathbf{W}^{(k)}(n)) \label{mdmeq2}
\end{eqnarray}
where $\boldsymbol{\theta}^{(k)}(n)^T = (\boldsymbol{\theta}_1^{(k)}(n)^T, \ldots, \boldsymbol{\theta}_P^{(k)}(n)^T)$ is the concatenated parameter vector and $V_i^{(k)}(n)$, $\bm{\Gamma}^{(k)}(n)$, $\bm{W}^{(k)}(n)$ must be specified.
The equations of the MDM can be viewed as a collection of nested univariate linear models, allowing the parameters to be estimated using well-known Kalman filter recurrences over time. 
We refer the reader to \cite{Costa} for further details regarding prior specification and computation of the local scores in this setting.

\begin{figure}[t]
\centering
\includegraphics[width = 0.8\textwidth,clip,trim = 3cm 13cm 16.5cm 2.5cm]{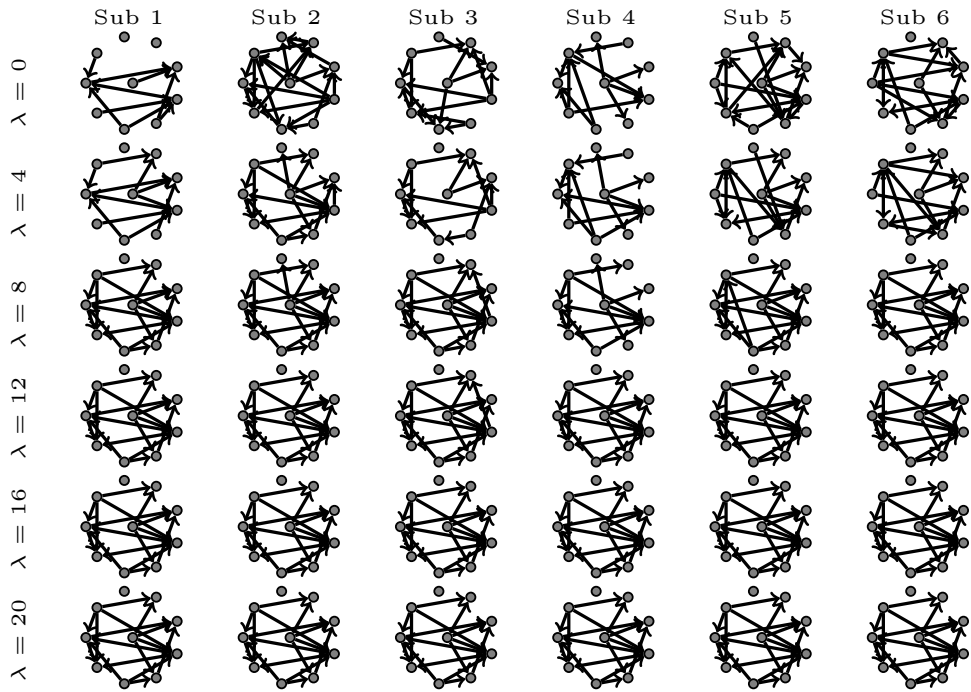}
\caption{Neuroscience data; exchangeable learning. [Here we simultaneously estimate subject-specific DAGs for different values of the regularity hyper-parameter $\lambda$.]}
\label{neuro1}
\end{figure}

{\bf Independent learning:} First, we applied independent estimation to each of the 6 subjects, with results shown in Fig. \ref{neuro1} (top). These results are unsatisfactory on neurological grounds since it is anticipated that connectivity does not change grossly between subjects.

{\bf Exchangeable learning:} Next, we inferred DAGs $G^{(1:K)}$ under an exchangeability assumption, for various values of the regularity hyper-parameter $\lambda$ (Fig. \ref{neuro1}), 
subjectively set to $\lambda = 4$.
(Alternative approaches to set $\lambda$  include AIC, as described above, or exploiting technical replicate data and examination of Bayes factors, as discussed at length in \cite{Oates5}).

\begin{figure}[t]
\centering
\includegraphics[width = \textwidth,clip,trim = 3.3cm 11.5cm 14.5cm 2.5cm]{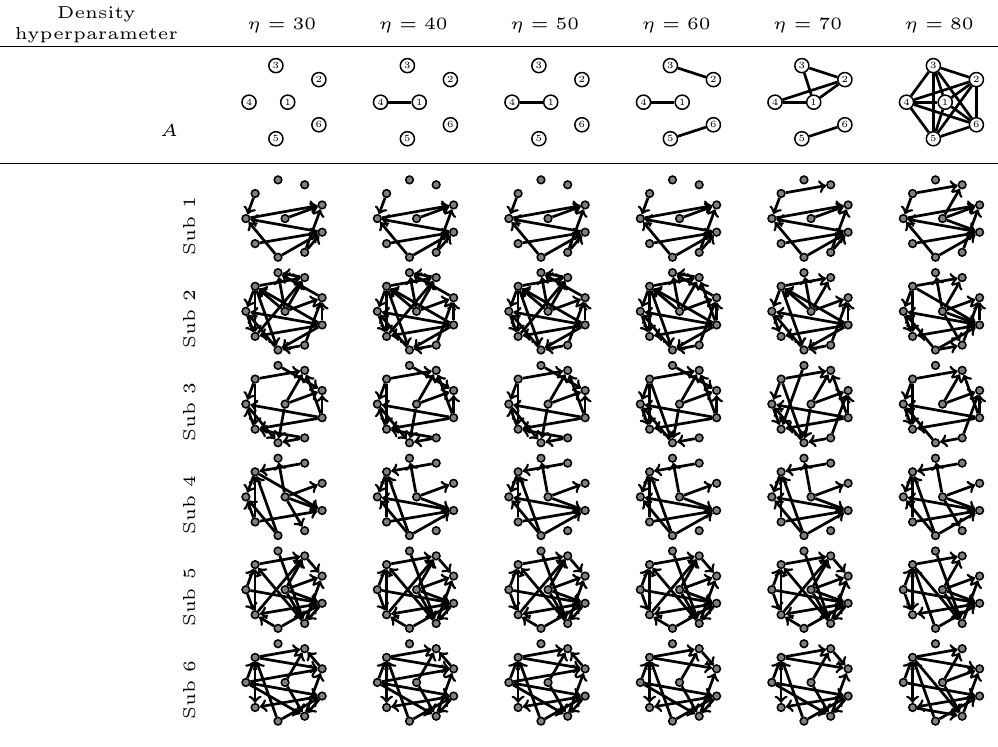}
\caption{Neuroscience data; non-exchangeable learning. [Here we simultaneously estimate both subject-specific DAGs and the network $A$ that relates subjects. The regularity hyper-parameter $\lambda = 4$ was fixed whilst the density hyper-parameter $\eta$ was varied.]}
\label{N neuro}
\end{figure}

{\bf Non-exchangeable learning:} Based on $\lambda=4$ we performed non-exchangeable learning in order to estimate similarities and differences between the 6 subjects in terms of their neural connectivity. 
Specifically, we estimate a network $A$ on these 6 subjects, based on various values of the density hyper-parameter $\eta$.
Results, shown in Fig. \ref{N neuro}, indicate that at $\eta = 60$, subjects 1 and 4, 2 and 3, and 5 and 6 exhibit the most similar graphical structures.
Non-edges $(k,l) \notin A$ in the network $A$ imply differences in the subjects' graphical structure; such information may therefore be used to identify subjects with particularly aberrant connectivity for further scientific investigation.
Going further, we see that subjects 5 and 6 are perhaps the most unusual, being totally disconnected from subjects 1 to 4 in the network $A$ when $\eta = 70$.
However, we recommend caution in interpreting $\hat{A}$, due to the difficulty in  estimation that we observed in simulations.
We note also that $\eta$ can also be elicited based on Bayes factors, as discussed in \cite{Oates5}.

Further analysis and interpretation of these results can be found in the companion paper \cite{Oates5}.
We emphasise that this analysis is exploratory and not confirmatory, but could be used to generate hypotheses for further experimental investigation.

\section{Discussion} \label{discuss}

This paper introduced a general statistical framework for joint estimation of multiple DAGs. We considered  regularisation based on graphical structure and showed how MAP estimators in this setting could be characterised as the solutions to ILPs that admit efficient exact algorithms based on branching-and-cutting as well as constraint propagation.
We believe these are the first exact algorithms for the general case of multiple general DAG models (i.e. with no restriction on ordering of the variables).
In our general framework for joint learning, we allowed also for dependence between the units themselves, that generalises previous exchangeable formulations.
Results obtained both from simulations and from an application in neuroscience demonstrate that joint estimation of DAGs can offer increased statistical efficiency relative to independent estimation.
However we observed that the relationships between the units themselves were  difficult to estimate from data.

In the neuroscience application we illustrated how sophisticated likelihood models such as MDMs can be used within our framework.
Importantly, our methodology requires only that local scores are pre-computed and cached; it is therefore possible to apply exact algorithms retrospectively, following  independent analyses of individual datasets, without the need to recompute local scores.
In practice we anticipate that the rationale and appropriateness of joint modelling will need be assessed by domain experts on a case-by-case basis.

Our computations were performed using the GOBNILP package
\citep{Bartlett}, that includes sophisticated routines for estimation of individual DAGs (see supplementary materials).
Nevertheless our simulation results for large numbers $P$ of variables
and $K$ of units required considerable computational time (in a worst-case scenario).  
It is becoming commonplace in many applications for the number of
units to become very large. 
It remains to be seen whether our approach can scale up to larger problems.
Recent work where GOBNILP was able to find
 MAP DAGs with 1614 nodes in between 3 and 42 minutes  are encouraging  \citep{Sheehan}.
In practice, as discussed in
\cite{Bartlett}, it is very hard to estimate the time required to
solve an ILP in advance; rather this depends in a
highly non-trivial way on the details of the problem and is related
to the ``phase-transition'' in SAT solving. 

We note that the results of this
  paper are based on the ILP framework of
  \cite{Jaakola} for individual DAGs.  An alternative ILP formulation,
  known as ``characteristic imsets'', was proposed by \cite{Studeny} specifically for BNs.
 Characteristic imsets are closely related to the essential graph of a BN
  \citep{Pearl} and have the property that
  Markov-equivalent BNs are score equivalent.  In contrast, the
  approaches we pursued involved a non-unique
  representation of the essential graph.  
  This flexibility is important when dealing with general DAG models including MDMs, that are uniquely identifiable from data, but are less natural in the context of BNs.
At present, the computational
  performance of characteristic imsets on individual BNs is inferior to the approach of \cite{Jaakola}
  pursued here \citep{Studeny2}.  
  In addition it is currently unclear how prior structural information might be incorporated into that framework.
  On a related note, it would be interesting to design a joint version of the well-known PC algorithm \citep{Spirtes} that complements score-based estimation.


Interesting extensions include:
\begin{enumerate}
\item {\bf Information sharing for parameters.} In many applications it is reasonable to assume similarity of parameter values $\bm{\theta}_{1:P}^{(k)}$ between units. At present this appears to be challenging to include within our framework and represents an area for further research.

\item {\bf Computation.} The applications discussed in this paper were performed on a single CPU and limited in computational intensity to approximately $P,K \leq 10$. Since many emerging datasets, not least in biomedical applications, contain many more variables $P$ and many more units $K$ than we considered here, it would be interesting to investigate strategies for parallel and/or approximate computation that scale better to such regimes.
In the other direction, our methodology applies to complex local scores that may be estimated numerically; it would be interesting to gain a theoretical understanding of how any uncertainty in the estimates for the local scores would impact on the MAP estimators described here.
\item {\bf New statistical models.} By fixing the network $A$ in our general framework we can obtain a number of interesting statistical models: (i) Taking $A$ to be a chain allows us to associate units with a temporal ordering and thereby model time-evolving DAG structures. (ii) Introducing auxiliary DAGs $G^{(K+1:K+L)}$ and enforcing $A$ to be bipartite on the partition of auxiliary/non-auxiliary DAGs produces a mixture model for DAGs, such that each latent DAG is a graphical summary of the various DAG structures within its cluster. (iii) Extending this idea, with $\lambda \rightarrow \infty$ we recover a mixture model for DAGs similar to \cite{Thiesson}. These possibilities are currently being explored, with preliminary results for mixture models presented in \cite{Oates5}.
Preliminary indications suggest that imposing additional restrictions on $A$ in this way may serve to improve the identification of higher-order structure.
More generally, it is straightforward to extend our methods to the case of decomposable undirected models by imposing additional constraints that rule out immoralities.
\end{enumerate}

{\bf Supplementary Materials:} The supplementary materials include additional a discussion of the simulated data and the actual code used to produce the results herein.

{\bf Acknowledgements:} CJO was supported by the Centre for Research in Statistical Methodology (CRiSM) EPSRC EP/D002060/1. JC was supported by the Medical Research Council (Project Grant G1002312). SM was supported by the UK Medical Research Council and is a recipient of a Royal Society Wolfson Research Merit Award.
The authors are grateful to Lilia Carneiro da Costa and Tom Nichols who collaborated in the analysis of fMRI data and to Mark Bartlett who provided technical support with GOBNILP.
The authors also thank Diane Oyen and several other colleagues who provided feedback on an earlier draft.

\FloatBarrier
\newpage
\pagenumbering{gobble}

\noindent {\bf Supplement to ``Exact Estimation of Multiple Directed Acyclic Graphs''}
Chris. J. Oates (E-mail: {\it c.oates@warwick.ac.uk}) and Jim Q. Smith (E-mail: {\it j.q.smith@warwick.ac.uk}), Department of Statistics, University of Warwick, Coventry, CV4 7AL, UK.
Sach Mukherjee (E-mail: {\it sach@mrc-bsu.cam.ac.uk}), MRC Biostatistics Unit and School of Clinical Medicine, University of Cambridge, CB2 0SR, UK.
James Cussens (E-mail: {\it james.cussens@cs.york.ac.uk}), Department of Computer Science and York Centre for Complex Systems Analysis, University of York, YO10 5GE, UK.

\section*{Supplementary Text}

\begin{figure}[h]
\centering
\subfloat[]{\includegraphics[width = 0.6\textwidth,clip,trim = 0cm 0cm 0cm 0.6cm]{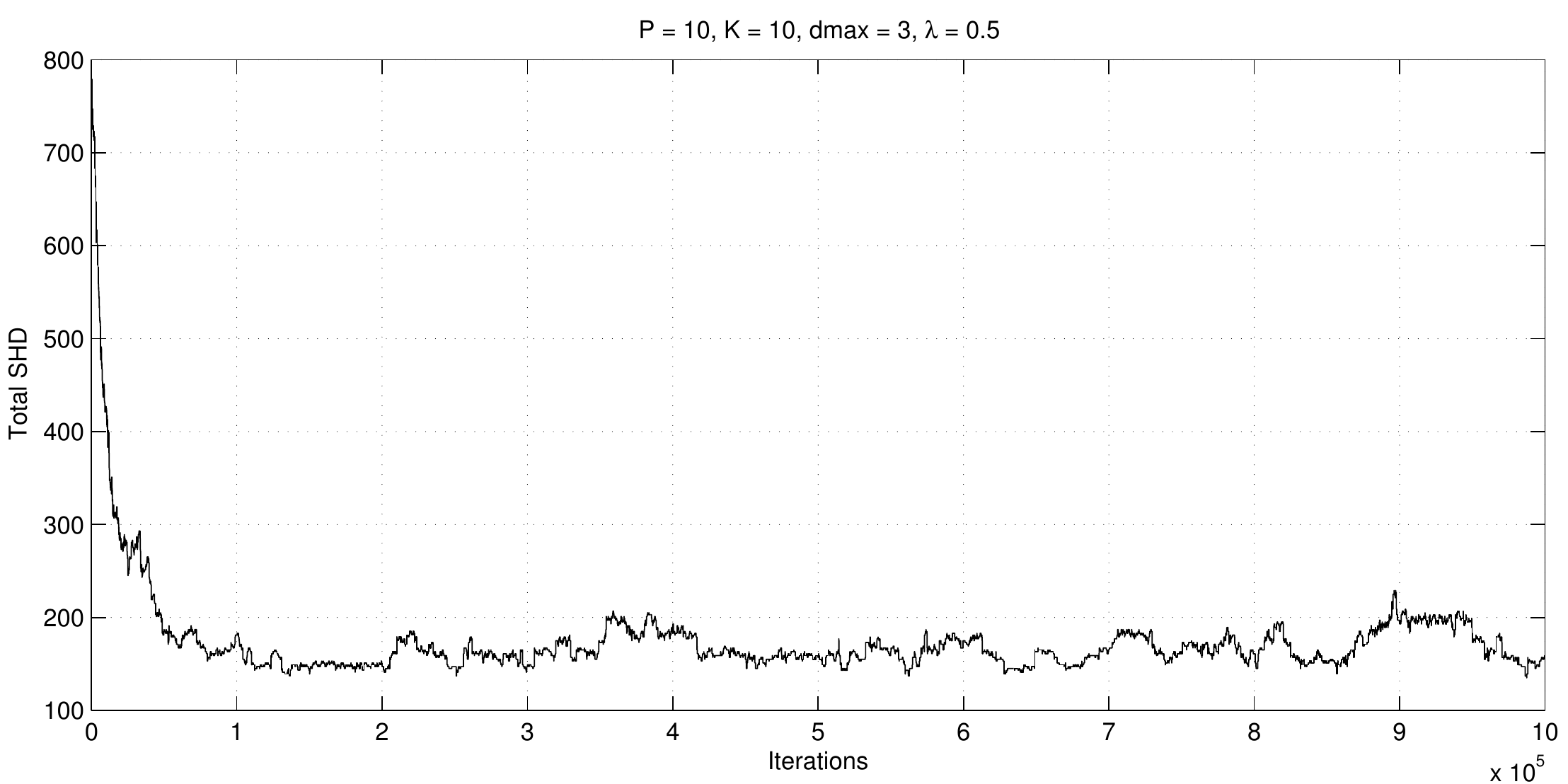} \label{trace}}

\subfloat[]{\includegraphics[width = 0.6\textwidth]{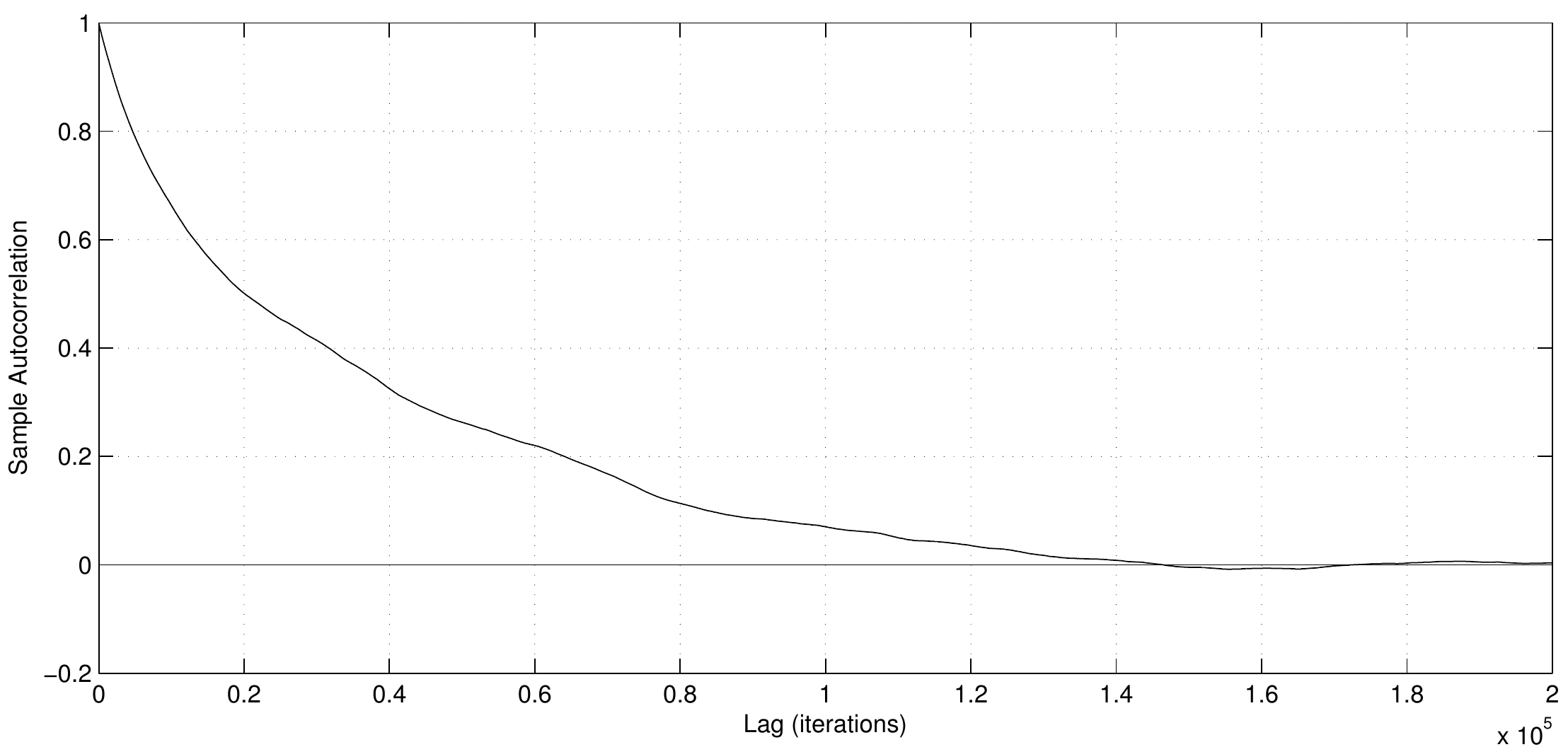} \label{ACF}}
\caption{Simulated data: Convergence diagnostics for sampling DAGs, based on $P=10$ variables and $K=10$ units. (a) MCMC trace plot for the total SHD between units. [Here $A$ is taken equal to the complete network.] (b) Autocorrelation function for the total SHD between units, corresponding to the trace plot of (a).}
\label{diagnose}
\end{figure}

In simulations we sampled the network $A$ directly from the prior $p(A)$ and then sampled DAGs $G^{(1:K)}$ from the prior conditional $p(G^{(1:K)}|A)$ using discrete-state-space MCMC.
Lines 2-12 in Algorithm \ref{MCMC} provide pseudo-code for the Metropolis-Hastings scheme that was employed.
Fig. \ref{diagnose} contains supporting convergence diagnostics.

As explained in the main text, the local likelihood $p(Y_i^{(k)}(n)|\bm{Y}_{G_i^{(k)}}(n),\bm{\theta}_i^{(k)}(n),G_i^{(k)})$ is arbitrary and its choice may influence the properties of the joint estimators that we wish to study.
We therefore took the most default approach of directly simulating the evidence terms $p(\bm{Y}_i^{(k)}|\bm{Y}_{G_i^{(k)}}^{(k)},G_i^{(k)})$ in Eqn. \ref{evidence} from log-normal distributions.
In this way we hope to obtain results that apply quite generally, without the potential for likelihood-specific anomalies.
For the AIC we assumed that $\dim(\bm{\theta}_i^{(k)}|G_i^{(k)}) = |G_i^{(k)}|$ so that each edge is associated with one free parameter as in, for example, a linear structural equation model.

Specifically, we appealed to the intuition that in many applications only a subset of models will be supported by data.
We therefore generated the log-evidence terms independently according to the mixture model $\frac{100-\alpha}{100} \times \delta(-\infty) + \frac{\alpha}{100} \times N(0,1)$, whilst the log-evidence of the true data-generating model was always simulated from $N(0,1)$. This ensures that the score $s^{(k)}(i,\pi)$ of the data-generating model belongs on average to the top $\alpha\%$ of all scores, whilst allowing us to discard $(1-\alpha)\%$ of all models and thereby reduce the computational burden in this simulation study.
All simulations were based on a value $\alpha = 15$.
To further mediate computational complexity we imposed an in-degree restriction $d_{\max} = 2$ for all simulation experiments; this could be relaxed at additional computational effort.
(Note that in the ``worst case'' simulations in the main text, all models were assigned log-scores from $N(0,1)$.)

In principle, suitable values for the hyper-parameters will scale with both $P$ and $K$.
To mitigate this effect and to allow us to employ the same candidate values $\lambda \in \{0,\frac{1}{2},1,2,\infty\}$, $\eta \in \{0,\frac{1}{2},1,2,\infty\}$ in each experiment, we rescaled the local scores $s^{(k)}(i,\pi)$ by a factor of $K^{-1}P^{-1}$.
Full pseudo-code for the simulation of the local scores $s^{(k)}(i,\pi)$ is provided in lines 13-24 of Algorithm \ref{MCMC}.

\begin{algorithm*}
\caption{Generation of simulated data. (Inputs: $P$ = number of variables; $K$ = number of units; $A$ = network; $d_{\max}$ = in-degree restriction)} \label{MCMC}
\begin{algorithmic}[1]
\Procedure{$[G^{(1:K)},s^{(1:K)}(1:P,\pi \subseteq \{1:P\})] =$ Simulate}{$P$,$K$,$A$,$d_{\max}$}
\State Initialise $G^{(1:K)}$.
\For{$i=1 : 20 \times P^2K^2$} \Comment{Use $20 \times P^2K^2$ Monte Carlo iterations.}
\State $k \sim \mathcal{U}(\{1:K\})$. \Comment{Select a unit $k$.}
\State $(j,i) \sim \mathcal{U}(\{1:P\} \times \{1:P\})$. \Comment{Select a pair $(j,i)$ of variables.}
\State $H^{(1:K)} \gets G^{(1:K)}$.
\State $H^{(k)}(j,i) \gets 1 - H^{(k)}(j,i)$. \Comment{Swap the status of the edge $(j,i)$ in unit $k$.}
\State $r \gets p(H^{(1:K)}|A) [ \text{isDAG}(H^{(k)}) ] [ |H_{i}^{(k)}| \leq d_{\max} ] / p(G^{(1:K)}|A)$.
\If{$\mathcal{U}(0,1) < r$} \Comment{Metropolis-Hastings accept/reject step.}
\State $G^{(1:K)} \gets H^{(1:K)}$.
\EndIf
\EndFor
\For{$k=1:K$}
\For{$i=1:P$}
\For{$\pi \subseteq \{1:P\} \setminus \{i\}$ s.t. $|\pi| \leq d_{\max}$}
\If{$\pi = G_i^{(k)}$}
\State $s^{(k)}(i,\pi) \sim N(0,1) - \log \binom{P}{|\pi|}$. \Comment{Simulate values for the local scores.}
\Else
\State $s^{(k)}(i,\pi) \sim \frac{100-\alpha}{100} \times \delta(-\infty) + \frac{\alpha}{100} \times N(0,1) - \log \binom{P}{|\pi|}$.
\EndIf
\State $s^{(k)}(i,\pi) \gets \frac{1}{KP} s^{(k)}(i,\pi)$. \Comment{Scale-adjust for hyper-parameters.}
\EndFor
\EndFor
\EndFor
\EndProcedure
\end{algorithmic}
\end{algorithm*}

\end{document}